\documentclass[10pt,conference,letterpaper]{IEEEtran}
\usepackage{times,amsmath,epsfig}
\usepackage{amssymb,amsthm,url}
\newtheorem{lemma}{Lemma}
\newtheorem{problem}{Problem}
\newtheorem{example}{Example}
\usepackage{balance}

\usepackage[textsize=footnotesize]{todonotes}
\presetkeys%
    {todonotes}%
    {inline,caption={}}{}

\newcommand{\partitle}[1]{\noindent {\em #1.}}

\title{Interactive Visual Data Exploration with Subjective Feedback: An Information-Theoretic Approach}

\author{%
{Kai Puolam\"{a}ki{\small $~^{\#,+}$}, Emilia Oikarinen{\small $~^{+}$}, Bo Kang{\small $~^{*}$},  Jefrey Lijffijt{\small $~^{*}$}, Tijl De Bie}{\small $~^{*}$}%
\vspace{1.6mm}\\
\fontsize{10}{10}\selectfont\itshape
$^{\#}$\,Department of Computer Science, Aalto University\\
Espoo, Finland\\
\fontsize{9}{9}\selectfont\ttfamily\upshape
\,kai.puolamaki@aalto.fi\\
\vspace{1.2mm}\\
\fontsize{10}{10}\selectfont\rmfamily\itshape
$^{+}$\,Finnish Institute of Occupational Health\\
Helsinki, Finland\\
\fontsize{9}{9}\selectfont\ttfamily\upshape
\,emilia.oikarinen@ttl.fi\\
\vspace{1.2mm}\\
\fontsize{10}{10}\selectfont\rmfamily\itshape
$^{*}$\,Department of Electronics and Information Systems, IDLab, Ghent University\\
Ghent, Belgium\\
\fontsize{9}{9}\selectfont\ttfamily\upshape
\,bo.kang@ugent.be, jefrey.lijffijt@ugent.be, tijl.debie@ugent.be%
}

\begin{document}
\maketitle

\begin{abstract}
%
  Visual exploration of high-dimensional real-valued datasets 
  is a fundamental task in exploratory data
  analysis (EDA).
  Existing methods use predefined criteria to
  choose the representation of data.
  There is a lack of methods that
  (i) elicit from the user what  she has learned from the data and
  (ii) show patterns that she does not know yet.
%
%
  We construct a theoretical model where identified patterns can be input as knowledge to the system.
  The knowledge syntax here is intuitive, such as ``this set of points forms a
  cluster'', and 
  requires no knowledge of maths.
  This background knowledge is used to find a Maximum Entropy distribution of the data, after which
  the system provides the user data projections in which the data
  and the Maximum Entropy distribution differ the most, hence showing the
  user aspects of the data that are maximally informative given the user's current knowledge.
%
  We provide an open source EDA system with
  tailored interactive visualizations to demonstrate these concepts.
  We study the 
  performance of the system and present use
  cases on both synthetic and real data.
%
%
  We find that the model and the prototype system allow the user to
  learn information efficiently from various data sources and  the
  system works sufficiently fast in practice.
%
%
  We conclude that the information theoretic approach to exploratory data
  analysis where patterns observed by a user are formalized as
  constraints provides a principled, intuitive, and efficient basis
  for constructing an  EDA system.
%
\end{abstract}



\section{Introduction}

\begin{figure*}[tp]
\centering
\includegraphics[width=\textwidth]{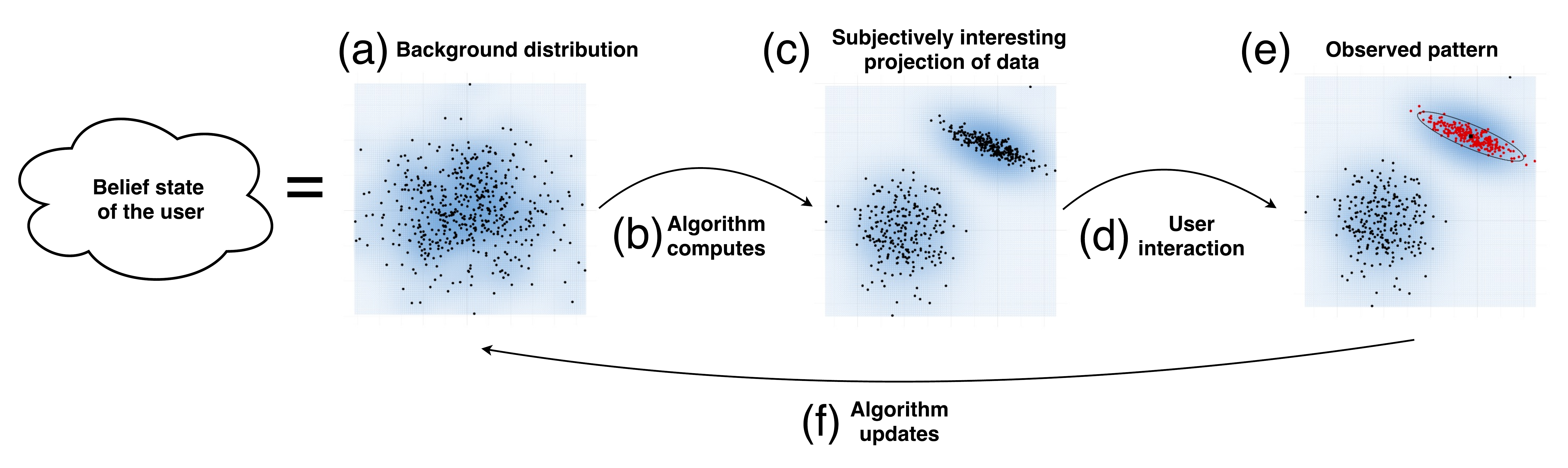}
\caption{Overview of the interaction process.\label{fig:overview}}
\end{figure*}

\begin{figure*}[tp]
\centering
\begin{tabular}{c@{}c@{}c}
\includegraphics[width=0.32\textwidth]{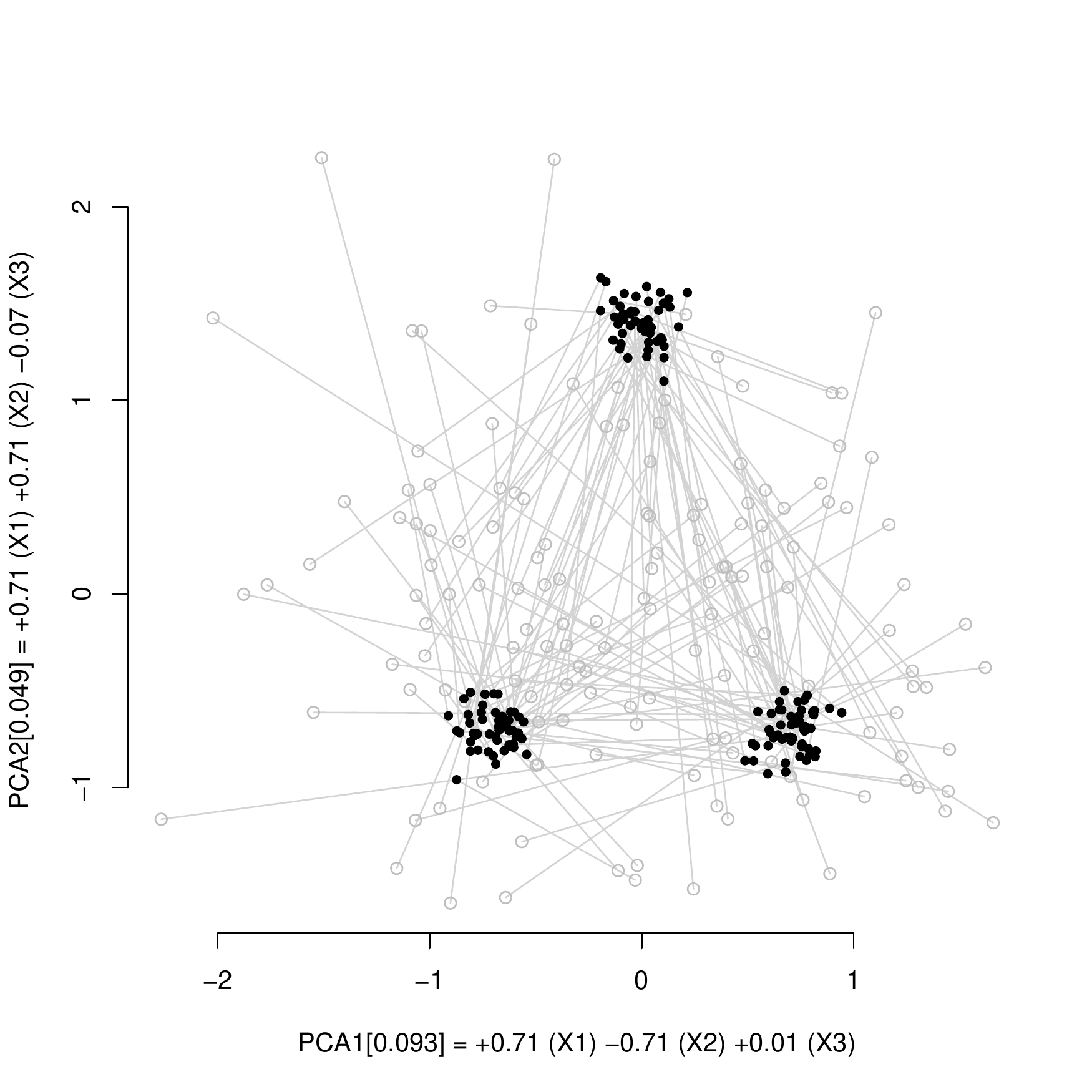} &
\includegraphics[width=0.32\textwidth]{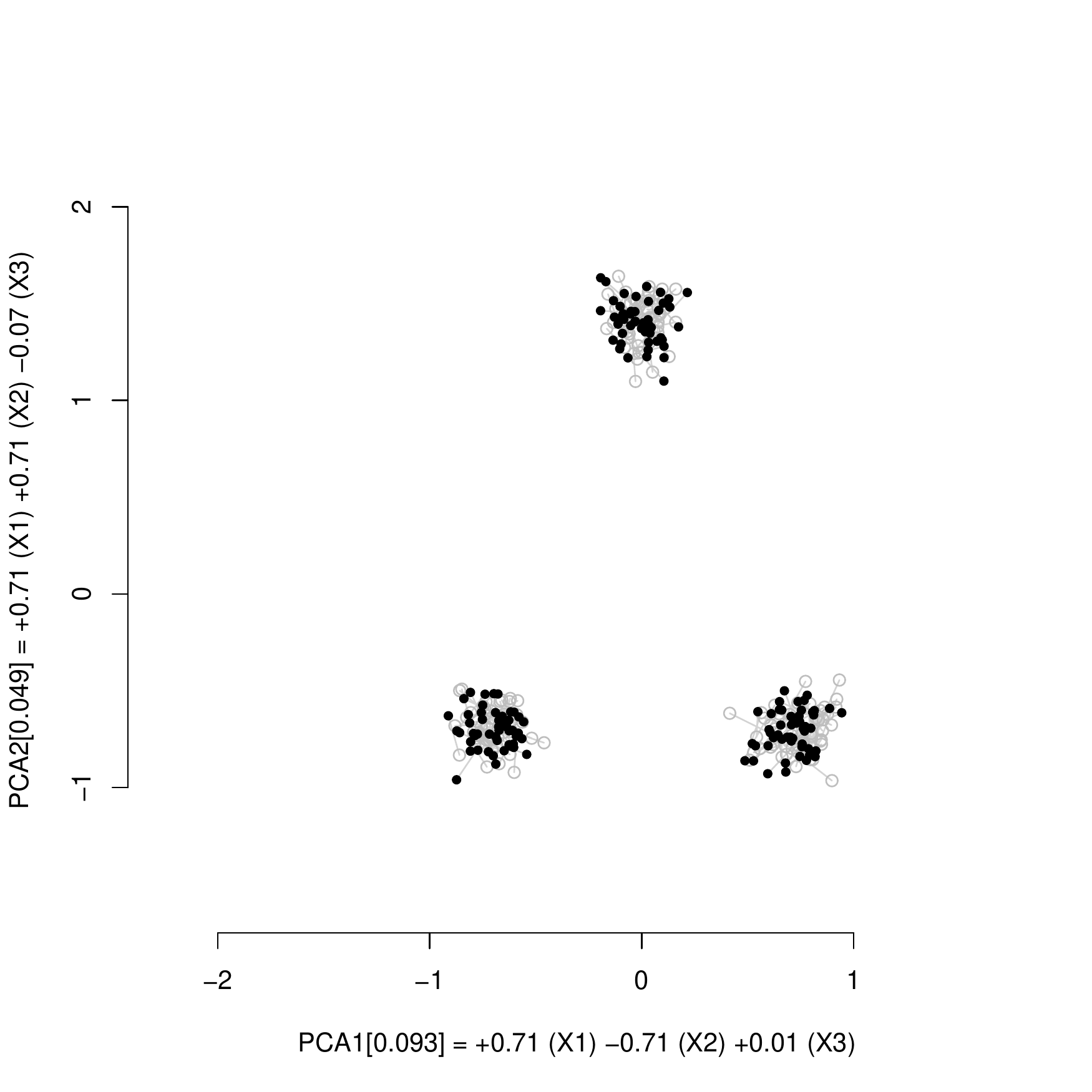} &
\includegraphics[width=0.32\textwidth]{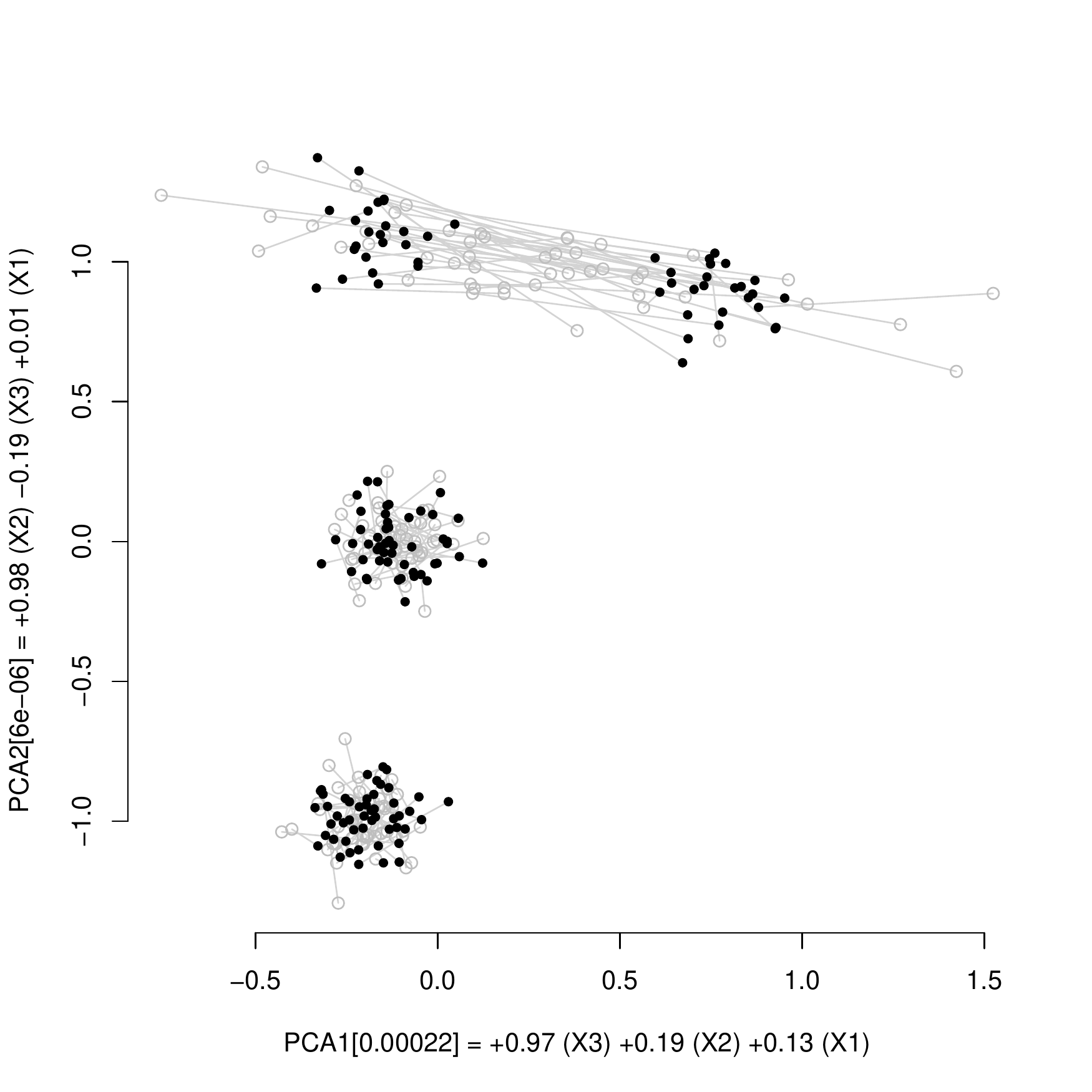} \\
\small{(a)} & \small{(b)} & \small{(c)} 
\end{tabular}
\caption{Synthetic data with 3 dimensions.  (a)  Projection of the data to the first two principal components together with a sample of background distribution;
(b) After the user's knowledge is taken into account, the updated background distribution matches the data in this projection;
(c) The user is then shown the next most informative projection.
\label{fig:intro_ex}}
\end{figure*}

Ever since Tukey's pioneering work on {\em exploratory data analysis} (EDA)
\cite{tukey1977exploratory}, 
effective exploration of data
has remained an art as much as a science. Indeed, while human
analysts are remarkably skilled in spotting patterns and relations in
adequately visualized data, coming up with insightful visualizations
is hard task to formalize, let alone to automate.

As a result, EDA systems require significant
expertise to use effectively. However, with the increasing
availability and importance of data, data analysts with sufficient
expertise are becoming a scarce resource. Thus, further research into
automating the search for insightful data visualizations is becoming
increasingly critical.

Modern computational methods for dimensionality reduction, such as
Projection Pursuit and manifold learning, allow one to spot complex
relations from the data automatically and to present them visually.
Their drawback is however that the criteria by which the views are found
are defined by static objective functions. The resulting visualizations may
or may not be informative for the user and task at hand. Often such
visualizations show the most prominent features of the data, while the
user might be interested in other subtler structures. It would therefore be
of a great help if the user could efficiently tell the system what she
already knows and the system could utilize this when deciding what to
show the user next. Achieving this is the main objective of this paper.

We present a novel interaction system based on solid theoretical principles.
The main idea of the system is shown
in Fig. \ref{fig:overview}. The computer maintains a distribution,
called the {\em background distribution} (\ref{fig:overview}a), modelling the
belief state of the user. The system shows 
the user projections
in which the data and the background distribution differ the most
(\ref{fig:overview}b,c). The user marks in the projection the patterns she
has observed (\ref{fig:overview}d,e) and the computer then uses these to
update the background distribution (\ref{fig:overview}f). The process is
iterated until the user is satisfied, i.e., typically when there are no 
notable differences between the data and the background distribution.

{\em Example.}
Specifically, the data considered in this work is a set of
$d$-dimensional ($d$-D) data points.  To illustrate the envisioned data
exploration process, we synthesized a 3-D dataset with 150
points such that there are two clusters
of 50 points and two of 25 points. The smaller clusters are
partially overlapping in the third dimension.  Looking at the first
two principal components, one can only observe three clusters with 50
points each (similarly to the black points in
Fig. \ref{fig:intro_ex}a).

In our interactive approach, the data analyst will learn not only that there are actually four clusters, but also that two of the four clusters correspond to a single cluster in the first view of the data.
The visualizations considered are scatter plots of the data points after projection onto a 2-D subspace, as in Projection Pursuit \cite{friedman1974,huber1985}.
The projection chosen for visualization is the one that (for a certain statistic) is maximally different with respect to the background distribution that represents the user's current understanding of the data.

In addition to showing the data in the scatterplot, we display a sample from the background distribution as gray points (and lines that connect the respective points, to give an indication of the displacement in the background distribution, per data point); see Fig. \ref{fig:intro_ex} for an example and Sec. \ref{sec:sider} for details. The data analyst's interaction consists of informing the system about sets of data points they perceive to form clusters within this scatter plot.
The system then takes the information about the user's knowledge of the data into account and updates the background distribution accordingly (Fig. \ref{fig:intro_ex}b).
%

When we have ascertained ourselves that the background distribution matches the data in the projection as we think it should, the system can be instructed to
 find another 2-D subspace to project the data onto. The new
 projection displayed is the one that is maximally insightful
 \emph{considering the updated background distribution}.  The next
 projection is shown in Fig. \ref{fig:intro_ex}c and reveals that one
 of the three clusters from the previous view can in fact be split
 into two. The user can now add further knowledge to the
 background distribution by selecting the two uppermost clusters and the process can be repeated.
 For our 3-D dataset, after the
 background distribution is updated upon addition of the new
 knowledge, the data and the background distribution match,
 and in this case, further projections will not reveal any additional structure.  

\begin{figure}[tp]
\includegraphics[width=0.95\columnwidth]{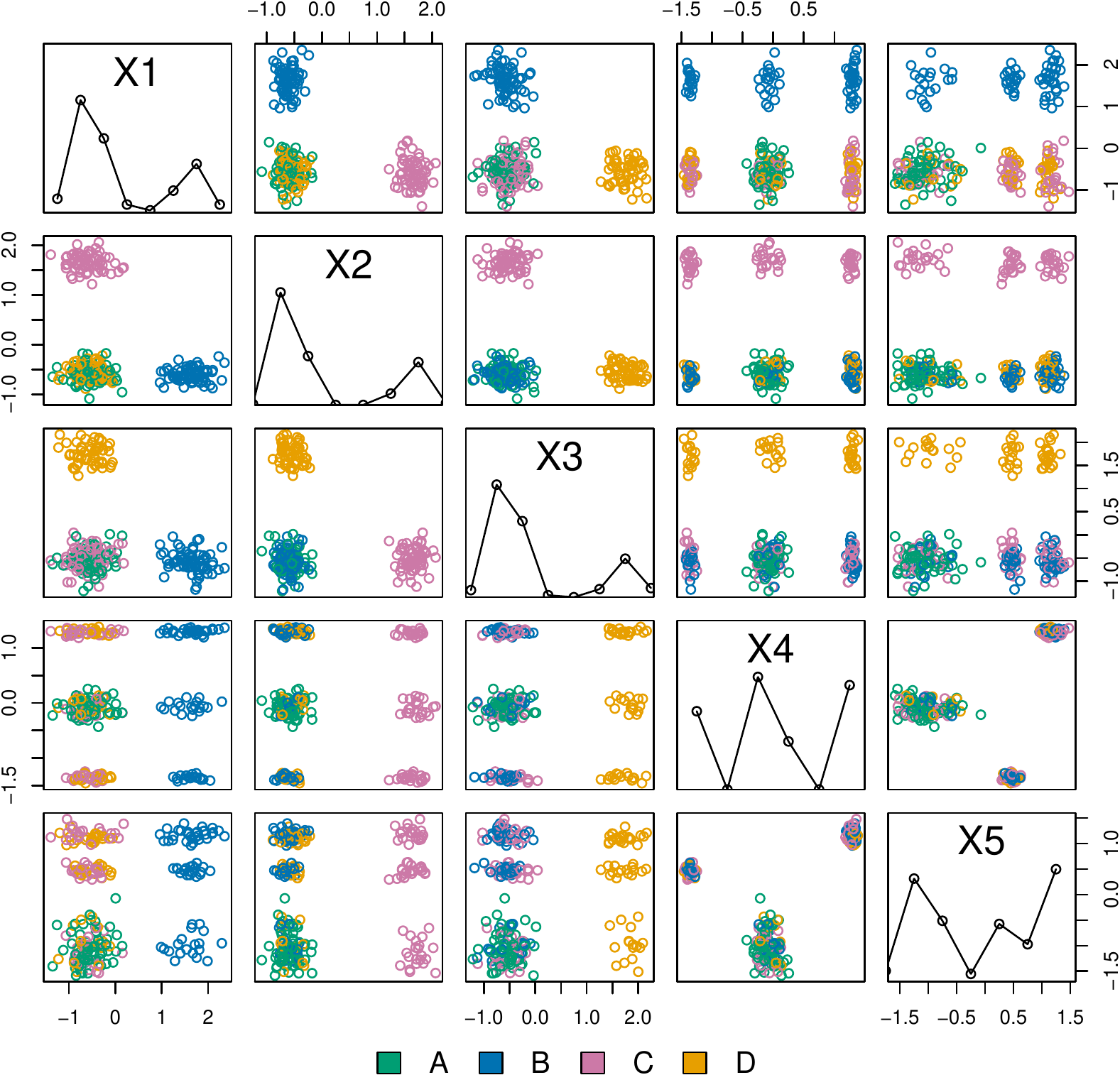} 
\caption{A pairplot of the synthetic data $\hat{\bf X}_{5}$. Colors correspond to the cluster identities $A, B, C$, and $D$ (the grouping that exists in the first three dimensions). Plot based on a sample of 250 points from the data.
 \label{fig:toy5}}
\end{figure}

\subsection{Contributions and outline of the paper}
The contributions of this paper are:
\begin{itemize}
  \item[-] We review how to formalize and efficiently find a background distribution accounting for a user's knowledge, in terms of a constrained Maximum Entropy distribution.
  \item[-] 
  A principled way to obtain projections 
    showing the maximal difference between the data and the background
    distribution for the PCA and ICA objectives, by \emph{whitening} the data with respect to the background distribution.
  \item[-] An interaction model by which the user can input what she has learned from the data, in terms of constraints.
  \item[-] An experimental evaluation of the computational performance of
    the method and use cases on real data.
  \item[-] A free open source application demonstrating the method.
\end{itemize}

The paper is structured as follows: we describe the method and
the algorithms in Sec. \ref{sec:method}, present the proof-of-concept open
source implementation {\sc sideR} in Sec. \ref{sec:sider}, report the
results of the experiments in Sec. \ref{sec:experiments}, discuss the
related work in Sec. \ref{sec:related}, and finally conclude in
Sec. \ref{sec:conclusion}.


\section{Methods}
\label{sec:method}

\subsection{Maximum Entropy background distribution}

\partitle{Preliminaries} The dataset consists of $n$ $d$-dimensional real vectors $\hat{\bf
  x}_i\in{\mathbb{R}}^d$, where $i\in[n]=\{1,\ldots,n\}$. The whole
dataset is represented by a real-valued matrix $\hat{\bf X}=\left(\hat{\bf
  x}_1\hat{\bf x}_2\ldots\hat{\bf
  x}_n\right)^T\in{\mathbb{R}}^{n\times d}$. We use hatted variables
(e.g., $\hat{\bf X}$) to denote the observed data and non-hatted
variables to denote the respective random variables (e.g., ${\bf X}$).

\vspace{2mm}\noindent{\bf Running example (Fig.~\ref{fig:toy5}).} {\em To illustrate the central concepts of the approach, 
we generated a synthetic dataset $\hat{\bf X}_{5}$ of 1000 data vectors in five dimensions.
The dataset is designed so that along dimensions 1--3 it can be clustered into four clusters (labeled $A,B,C,D$)
and along dimensions 4--5 into three clusters (labeled $E,F,G$). The clusters in dimensions 1--3 are located such that in any 2-D projection along these dimensions cluster $A$ overlaps with one of the clusters $B,C$, or $D$. 
The cluster structure in dimensions 4--5 is loosely related to the cluster structure in dimensions 1--3:
with 75\% probability a data vector belonging to clusters $B,C$, or $D$  belongs to one of clusters $E$ and $F$.
The remaining points belong to cluster $G$.
The pairplot in  Fig.~\ref{fig:toy5} shows the structure of the data (colors correspond to the cluster identities A, B, C, D).}

\vspace{2mm}
\partitle{Constraints} To implement the envisioned interaction scheme, we wish to define constraints on the data and to
find a Maximum Entropy distribution such that the constraints set by the
user are satisfied. Intuitively, the more constraints we have, the closer
the distribution should be to the true data.

We must define \emph{some} initial background distribution. Reasonable and convenient is to assume that the initial background distribution equals a spherical Gaussian distribution with zero mean and unit variance, given by
\begin{equation}\label{eq:q}
  q({\bf X})\propto\exp{\left(-\sum\nolimits_{i=1}^n{{\bf x}_i^T{\bf
        x}_i}/2\right)}.
\end{equation}

\begin{figure*}[tp]
\begin{center}
\begin{tabular}{cc}
\includegraphics[width=0.8\columnwidth]{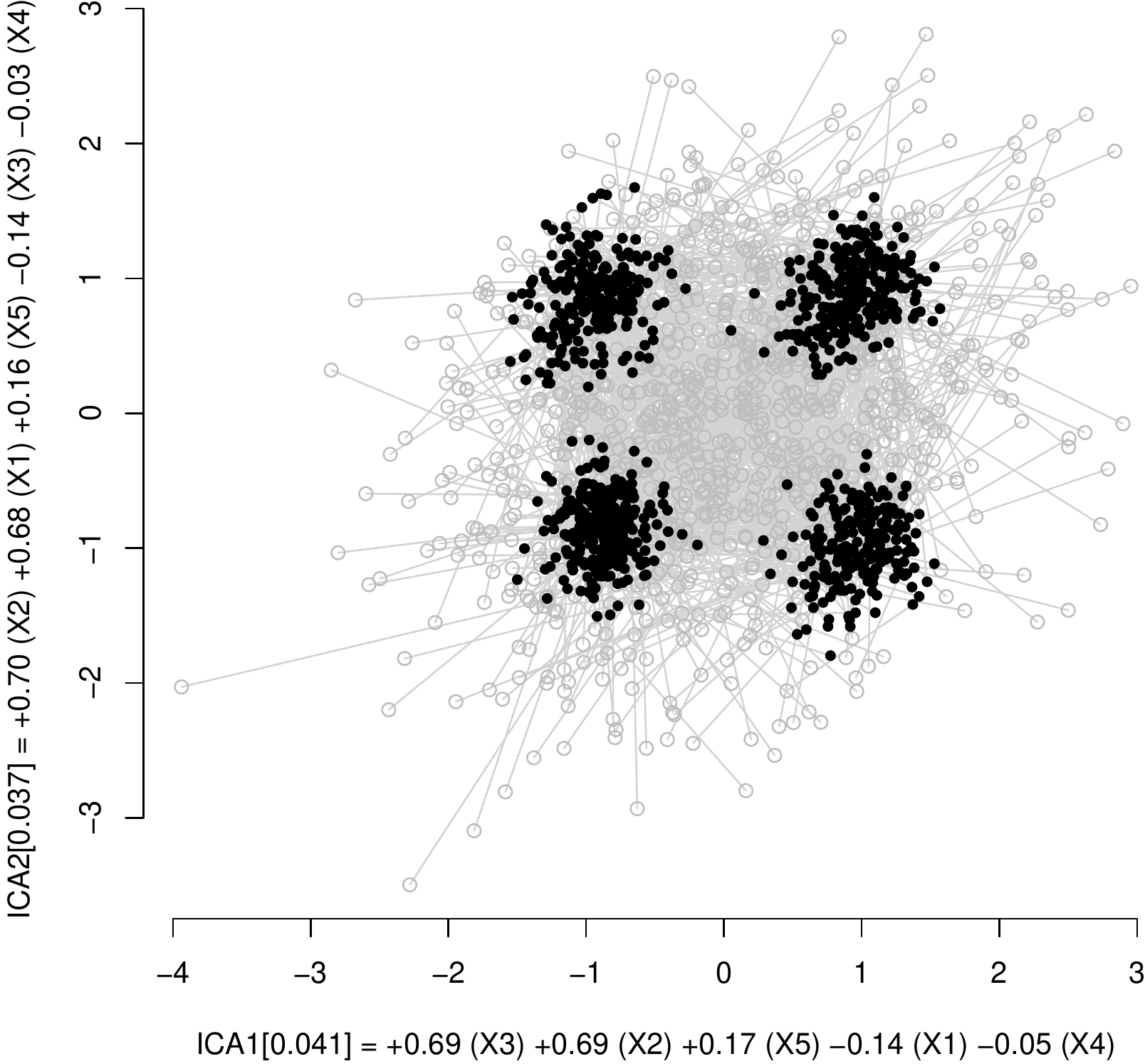} &
\includegraphics[width=0.8\columnwidth]{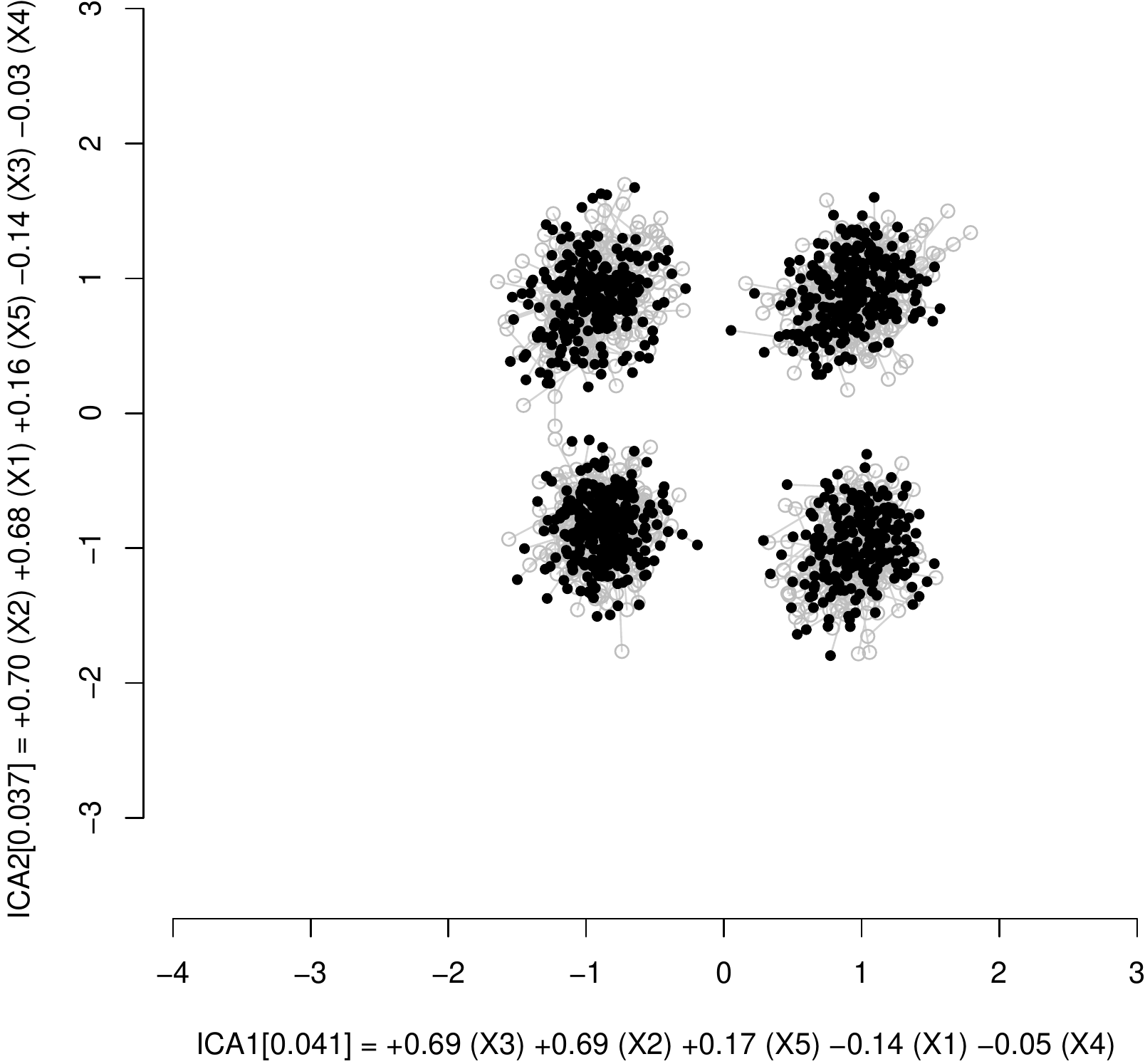} \\
\small{(a)} & \small{(b)} \\
\includegraphics[width=0.8\columnwidth]{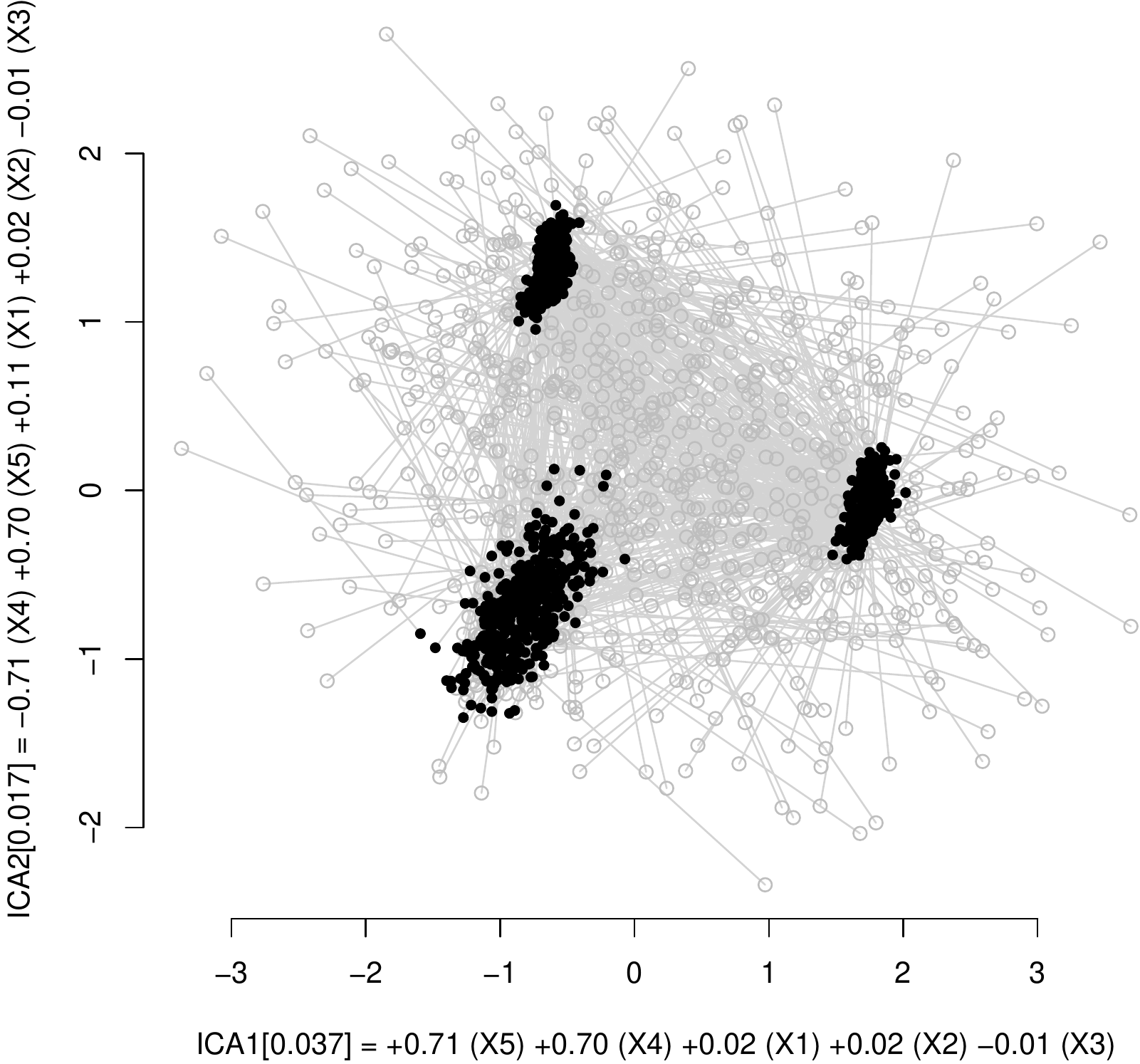} &
\includegraphics[width=0.8\columnwidth]{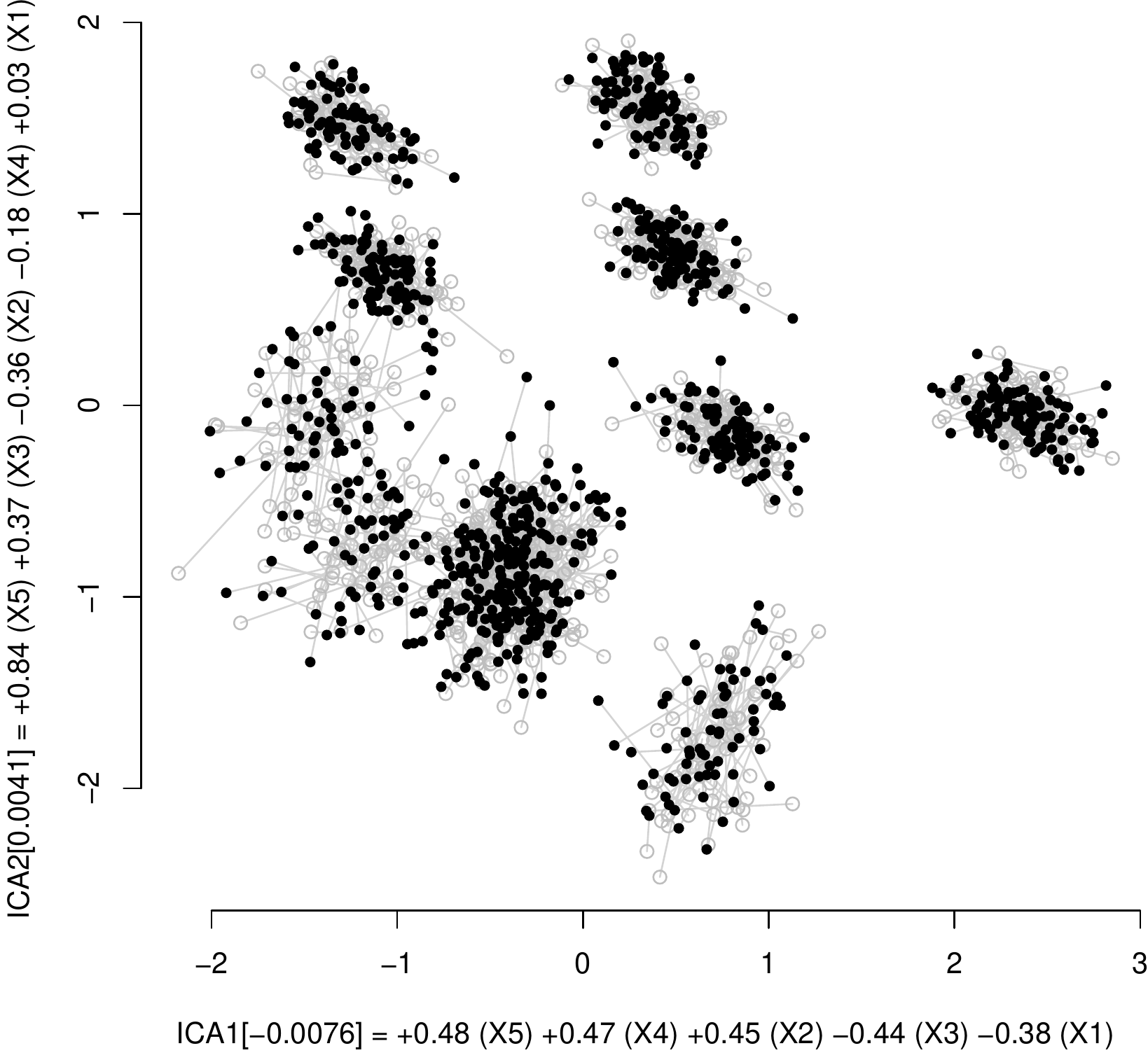}
\\ \small{(c)} & \small{(d)}
\end{tabular}
\caption{(a) The synthetic data  $\hat{\bf X}_{5}$ projected into the first two ICA components and shown together with a sample of the background distribution (in gray). With no knowledge about the belief state of the user, the background distribution is Gaussian. (b) The same projection with the background distribution updated to take into account cluster constraints for the four visible clusters.
(c) The next most informative ICA projection for  $\hat{\bf X}_{5}$.
(d)~The ICA projection obtained after  further cluster constraints for the three visibile clusters have been added and the background distribution has been updated.
 \label{fig:toy5_ica0}}
\end{center}
\end{figure*}

\begin{example}
As illustrated in Fig. \ref{fig:overview}, the interaction process is such that the user is shown 2-D projections (\ref{fig:overview}c) where
the data and the background distribution differ the most.
Initially, when there is no knowledge about the belief state of the user, the background distribution (\ref{fig:overview}a) is assumed to be
a spherical Gaussian distribution with zero mean and unit variance.
The initial view shown to the user is a projection of the data onto the first two PCA or ICA components.
Fig. \ref{fig:toy5_ica0}a shows the projection of $\hat{\bf X}_{5}$ onto the first two ICA components.
One can observe the cluster structure in the first three dimensions.
The gray points represent a sample from the background distribution.  When shown together with the data, it becomes evident that the data and the background distribution differ.
\end{example}

Subsequently, we can
define constraints on subsets of points in
${\mathbb{R}}^{n\times d}$ for a given projection by introducing
{\em linear and quadratic constraint functions} \cite{lijffijt_forthcoming}.
A constraint is
parametrized by the subset of rows $I\subseteq [n]$
that are involved and a projection vector $\mathbf{w}\in \mathbb{R}^d$.
The {\em
  linear constraint function} is defined by
\begin{equation}\label{eq:flin}
f_{\textrm{lin}}(\mathbf{X},I,\mathbf{w})=\sum\nolimits_{i\in I}{\mathbf{w}^T\mathbf{x}_i},
\end{equation}
and the {\em quadratic constraint function} by
\begin{equation}\label{eq:fquad}
f_{\textrm{quad}}(\mathbf{X},I,\mathbf{w})=\sum\nolimits_{i\in I}{\left(\mathbf{w}^T\left(\mathbf{
    x}_i-\mathbf{\hat m}_I\right)\right)^2},
\end{equation}
where we have used
\begin{equation}\label{eq:mhat}
  \mathbf{\hat m}_I=\sum\nolimits_{i\in I}{\mathbf{\hat x}_i}/|I|.
\end{equation}

Notice that $\mathbf{\hat m}_I$ is not a random variable but a
constant that depends on the observed data. If it were a random
variable it would introduce cross-terms between rows and the
distribution would no longer be independent for different rows. In principle, we could set $\mathbf{\hat m}$ to any constant value, including zero.
However, for the numerical algorithm to converge quickly we use the
value specified by Eq.~\eqref{eq:mhat}.

We can use the linear and quadratic constraint functions to
define several types of knowledge a user may have about the data,
which we can then encode into the Maximum Entropy background distribution.

We can encode the mean and variance, i.e., the first and second moment of the marginal distribution, of each attribute:
\begin{itemize}
\item {\em Margin constraint} consists of a linear and a quadratic
  constraint for each of the columns in $[d]$, respectively, the total
  number of constraints being $2d$.
\end{itemize}
We can encode the mean and (co)variance statistics of a point cluster for all attributes:
\begin{itemize}
\item {\em Cluster constraint} is defined as follows.  We make a singular
  value decomposition (SVD) of the points in the cluster defined by
  $I$. Then a linear and a quadratic constraint is defined for each of
  the eigenvectors. This results in $2d$ constraints per cluster.
\end{itemize}
We can encode the mean and (co)variance statistics of the full data for all attributes:
\begin{itemize}
\item {\em 1-cluster constraint} is a special case of a cluster constraint
  where the full dataset is assumed to be in one single cluster (i.e.,
  $I=[n])$. Essentially, this means that the data is modeled by its
  principal components and the correlations are taken into account,
  unlike with the marginal constraints,  again resulting to $2d$
  constraints.
\end{itemize}
We can encode the mean and variance of a point cluster or the full data as shown in the current 2-D projection:
\begin{itemize}
\item {\em 2-D constraint} consists of  a linear and a quadratic
  constraint for the two vectors spanning the 2-D projection in
  question, resulting to $4$ constraints.
\end{itemize}

\partitle{The background distribution}
We denote a constraint by a triplet $C=(c,I,\mathbf{w})$, where
$c\in\{lin,quad\}$, and the constraint function is then given by
$f_c(\mathbf{ X},I,\mathbf{ w})$.  Our main problem, i.e., how to
update the background distribution given a set of constraints, can be
stated as follows.
\begin{problem}\label{pro:maxent}
  Given a dataset $\mathbf{\hat X}$ and
  $k$ constraints $\mathcal{C}=\{C^1,\ldots,C^k\}$, find a probability
  density $p$ over datasets ${\bf X}\in{\mathbb{R}}^{n\times d}$ such
  that the entropy defined by
  \begin{equation}\label{eq:maxent}
    S=-E_{p({\bf X})}\left[\log{\left(p({\bf X})/q({\bf X})\right)}\right]
  \end{equation}
  is maximized, while the following constraints are satisfied for all
  $t\in[k]$,
  \begin{equation}\label{eq:constraints}
    E_{p(\mathbf{ X})}\left[f_{c^t}(\mathbf{ X},I^t,\mathbf{ w}^t)\right]=\hat v^t,
  \end{equation}
  where $\hat v^t= f_{c^t}(\mathbf{ \hat X},I^t,\mathbf{ w}^t)$.
\end{problem}
We call the distribution $p$ that is a solution to the
Prob.~\ref{pro:maxent} the {\em background distribution}. Intuitively,
the background distribution is the maximally random distribution such
that the constraints are
preserved in expectation.  The form of the solution to
Prob. \ref{pro:maxent} is given by the following lemma.
\begin{lemma}
  The probability density $p$ that is a solution to
  Prob. \ref{pro:maxent} is of the form
  \begin{equation}\label{eq:lambdaeq}
    p({\bf X})\propto q({\bf
      X})\times\exp{\left(\sum\nolimits_{t=1}^k{\lambda^tf_{c^t}({\bf
          X},I^t,{\bf w}^t)}\right)},
  \end{equation}
  where $\lambda^t\in{\mathbb{R}}$ are real-valued parameters.
\end{lemma}
\noindent See, e.g., Ch.  6 of \cite{Cover05} for a proof.

We make an observation that adding a margin constraint or 1-cluster
constraint to the background distribution is equivalent to first
transforming the data to zero mean and unit variance or whitening of
the data, respectively.

Eq. \eqref{eq:lambdaeq} can also be written in
the form
\begin{equation}\label{eq:theta}
  p({\bf X}\mid\theta)\propto\exp{\left(-\sum\nolimits_{i=1}^n{
      \left({\bf x}_i-{\bf m}_i\right)^T
      \Sigma_i^{-1}
      \left({\bf x}_i-{\bf m}_i\right)/2}
    \right)},
\end{equation}
where the natural parameters are collectively denoted by $\theta
=\left\{\theta_i\right\}_{i\in[n]}
=\left\{\left(\Sigma^{-1}_i{\bf m}_i,\Sigma_i^{-1}\right)\right\}_{i\in[n]}$,
which can be written as sums of the terms of the form
$\lambda^tf_{c^t}({\bf X},I^t,{\bf w}^t)$ by matching the terms linear
and quadratic in ${\bf x}_i$ in Eqs. \eqref{eq:lambdaeq} and
\eqref{eq:theta}.  The dual parameters are given by
$\mu=\left\{\mu_i\right\}_{i\in[n]}= \left\{\left({\bf
  m}_i,\Sigma_i\right)\right\}_{i\in[n]}$ and can be obtained
from the natural parameters by simple matrix operations (see below).

Prob. \ref{pro:maxent} can be solved numerically as
follows. Initially, we set the lambda parameters to $\lambda_1=\ldots=\lambda_k=0$, with the natural dual parameters then
given by $\theta_i=\mu_i=\left({\bf 0},{\bf 1}\right)$ for all
$i\in[n]$. We then update the lambda parameters iteratively
as follows. Given some values for the lambda parameters and the
respective natural and dual parameters, we choose a constraint
$t\in[k]$ and find a value for $\lambda^t$ such that the constraint of
Eq. \eqref{eq:constraints} is satisfied for this chosen $t$. We then
iterate this process for all constraints $t\in[k]$ until
convergence. Due to convexity of the problem we are always guaranteed
to eventually end up in a globally optimal solution.

For a given set of lambda parameters we can find the natural
parameters in $\theta$ by a simple addition. The dual parameters can
be obtained from the natural parameters, e.g., by using matrix
inversion and multiplication operations. The expectation in
Eq. \eqref{eq:constraints} can be computed by using the dual
parameters and identities $E_{p({\bf X}\mid\theta)}\left[{\bf x_i}{\bf
    x_i}^T\right]=\Sigma_i+{\bf m}_i{\bf m}_i^T$ and $E_{p({\bf
    X}\mid\theta)}\left[{\bf x}_i\right]={\bf m}_i$.

\begin{example}
After observing the view in Fig. \ref{fig:toy5_ica0}a the user can  add a cluster constraint for each of the four clusters visible in the view. The background distribution is then updated to take into account the added constraints by solving Prob. \ref{pro:maxent}.
In Fig. \ref{fig:toy5_ica0}b a sample of the updated background distribution (gray points) is shown together with the data (black points).
\end{example}

\begin{figure*}[tp]
  \centering
  \begin{tabular}[b]{c}
    \includegraphics[width=0.67\columnwidth]{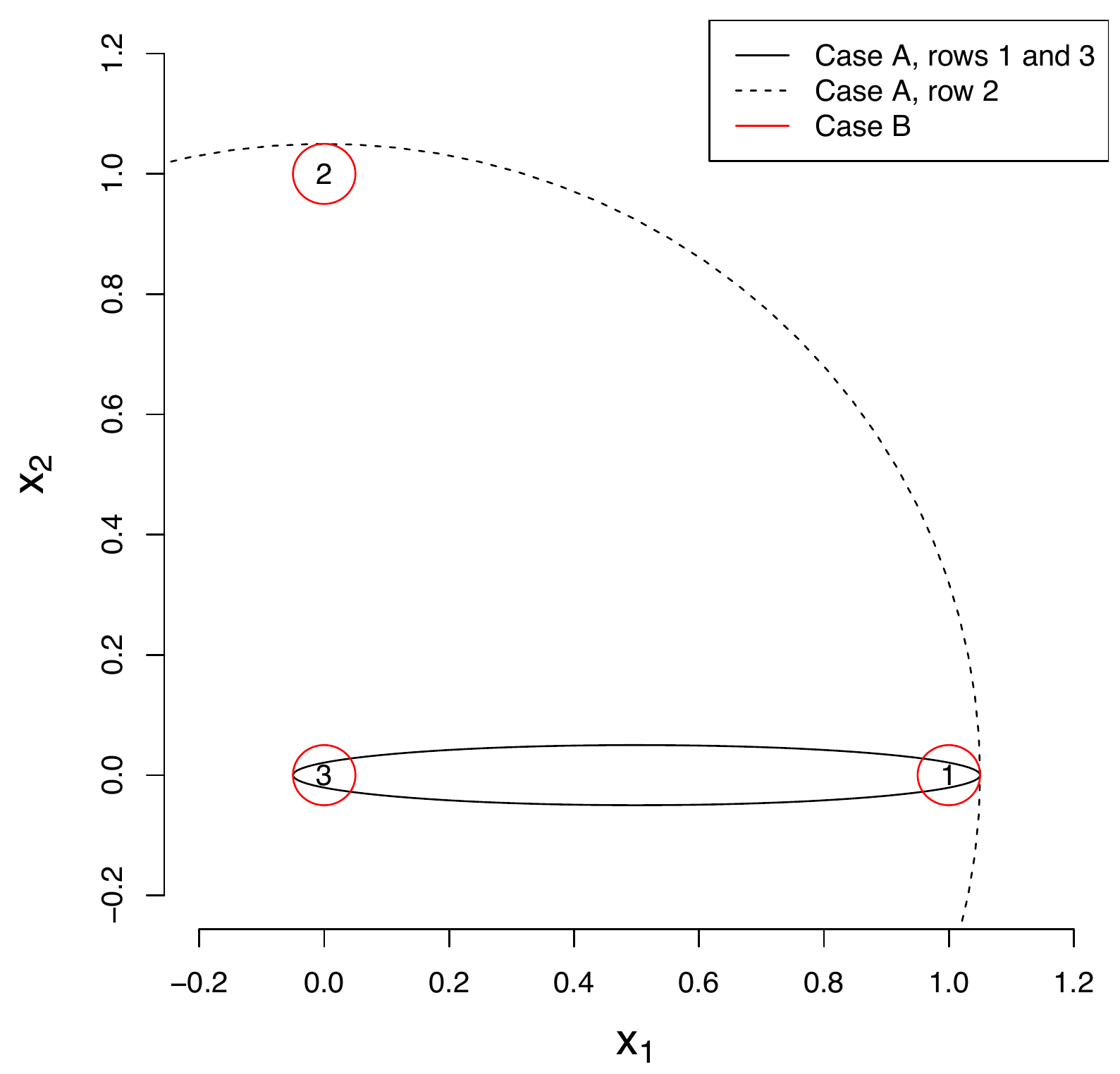}\\
    \small{(a)}
  \end{tabular}
  \hfil
  \begin{tabular}[b]{c}
    \includegraphics[width=0.67\columnwidth]{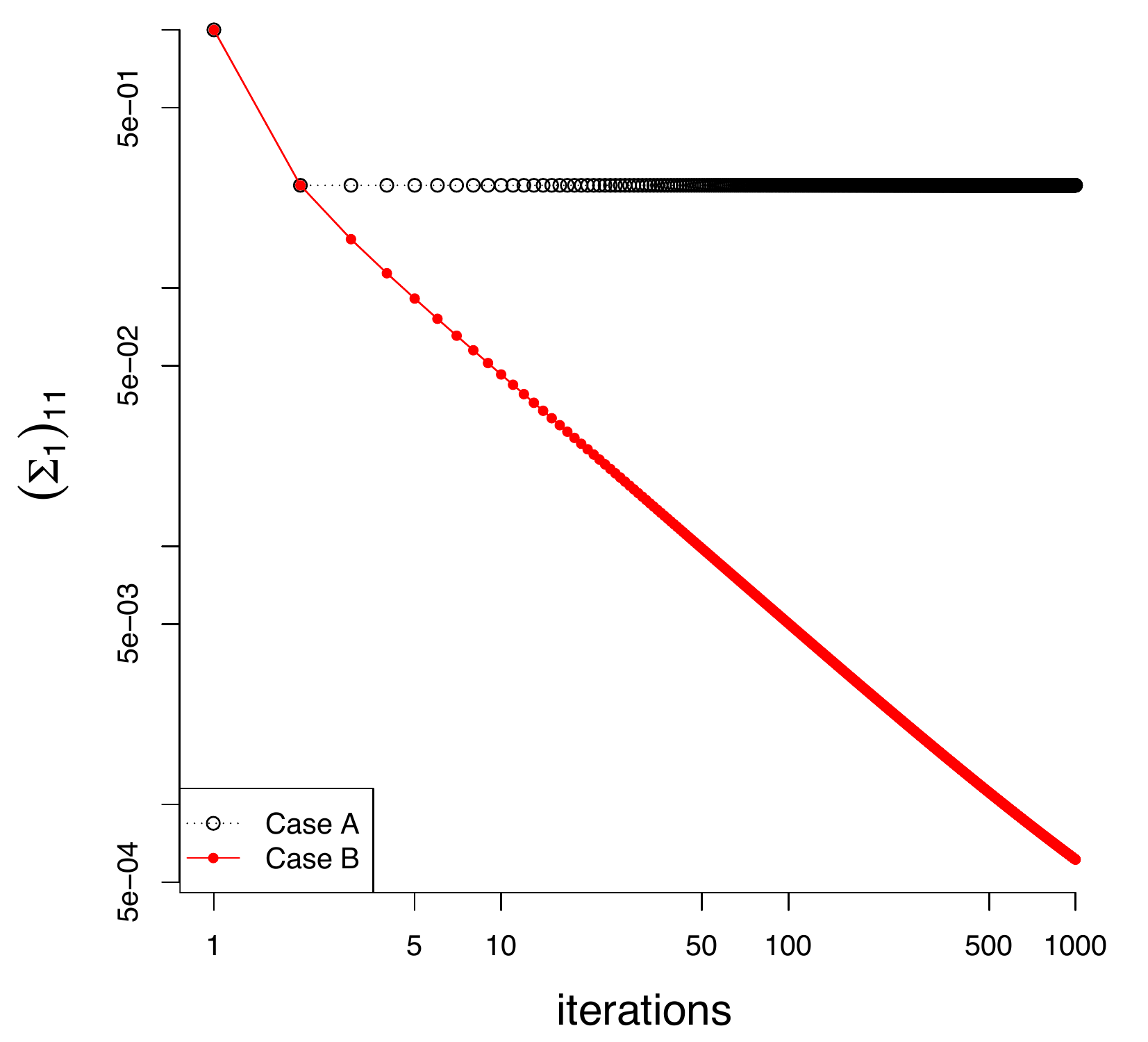}\\
    \small{(b)}
  \end{tabular}
\caption{\label{fig:conv}(a) Adversarial toy data and $1\sigma$
  confidence ellipsoids (expanded a bit for visual clarity in zero
  variance directions) for the background distribution. (b)
  Convergence of $(\Sigma_1)_{11}$ for two sets of constraints ${\mathcal C}_A$
  (black line) and ${\mathcal C}_B$ (red line).}
\end{figure*}

The straightforward implementation of the above mentioned optimization
process is inefficient because we need to store parameters for
$n$ rows and and because the matrix inversion is an $O(d^3)$ operation,
resulting to a time complexity of $O(nd^3)$.

We can
substantially speed up the computations by two observations. First,
two rows affected by the same constraints share equal
parameters. We need only to store and compute values of the
parameters $\theta_i$ and $\mu_i$ for ``equivalence classes'' of rows,
whose number depends on the number and overlap of constraints, but
not on $n$. Second, if we store both the natural and dual
parameters at each iteration, the constraint at an iteration
corresponds to a rank-1 update to the covariance matrix
$\Sigma^{-1}_i$. We can then use the Woodbury Matrix Identity
taking $O(d^2)$ time to compute the inverse, instead of $O(d^3)$.

Moreover, by storing the natural
and dual parameters at each step we do not need to explicitly store
the values of the lambda parameters. At each iteration we are only
interested in the change of $\lambda^t$ instead of its
absolute value. After these speedups we expect the optimization
to take $O(d^2)$ time per constraint and be asymptotically
independent of $n$. For simplicity, in the descriptions 
we retain the sums of the form $\sum\nolimits_{i\in I^t}{}$. 
In the implementation we replace these by the more efficient weighted
sums over the equivalence classes of rows.

\begin{table}[t]
\centering
    \caption{\label{tab:icascores} ICA scores (sorted with absolute value) for each of the iterative steps in Fig. \ref{fig:toy5_ica0}.
     }
\begin{tabular}{l|ccccc}
Projection & \multicolumn{5}{c}{ICA scores} \\
\hline
Fig. \ref{fig:toy5_ica0}a,b & 0.041  & 0.037 & 0.035  & 0.034 & -0.015 \\
Fig. \ref{fig:toy5_ica0}c  &  0.037  & 0.017 & 0.004 & -0.003 & -0.002 \\
Fig. \ref{fig:toy5_ica0}d & -0.008 & 0.004 & -0.003 & 0.003 & -0.002
\end{tabular}
  \end{table}
  
\subsubsection{Update rules}
We end up with the following update rules for the linear and quadratic
constraints, respectively, which we iterate until convergence. 
To simplify and clarify the notation we use parameters with tilde (e.g.,
$\tilde\Sigma$) to denote them before the update and
 parameters without (e.g., $\Sigma$) to denote the values after the update. We
further use the following short-hands, where $\lambda$ is the change in
$\lambda^t$: ${\bf b}_i=\tilde\Sigma_i\tilde\theta_{i1}$, $c_i={\bf
  b}_i^T{\bf w}^t$, $\Lambda_i=\lambda/\left(1+\lambda c_i\right)$,
$d_i={\bf b}_i^T\tilde\theta_{i1}$, $e_i=\tilde{\bf m}_i^T{\bf w}^t$,
$\delta=\hat{\bf m}_{I^t}^T{\bf w}^t$, $f_i=\lambda
\delta-\Lambda_id_i-\Lambda_i\lambda \delta c_i$, and
${\bf g}_i=\tilde\Sigma_i{\bf w}^t$.

For a {\bf linear constraint} $t$ the expectation is given by $v^t=
E_{p({\bf X}\mid\theta)}\left[ f_{lin}(\mathbf{ X},I^t,\mathbf{ w}^t)
  \right] =\sum\nolimits_{i\in I^t}{{\bf w}^{tT}{\bf m}_i}$. The
update rules for the parameters are given by
$\theta_{i1}=\tilde\theta_{i1}+\lambda{\bf w}^t$ and
$\mu_{i1}=\Sigma_i\theta_{i1}$.  Solving for $v^t=\hat v^t$
gives the required change in $\lambda^t$ as
\begin{equation}
  \lambda=\left(\hat v^t-\tilde v^t\right)/\left(
  \sum\nolimits_{i\in I^t}{{\bf w}^{tT}\tilde\Sigma_i{\bf w}^t}
  \right),
\end{equation}
where $\tilde v^t$ denotes the value of $v^t$ before the
update. Notice the change in $\lambda^t$ is zero if $\tilde v^t=\hat
v^t$, as expected.

For a {\bf quadratic constraint} $t$ the expectation is given by $v^t=
E_{p(\mathbf{X}\mid\theta)}\left[f_{quad}(\mathbf{ X},I^t,\mathbf{
    w}^t)\right]$ or $v^t={\bf w}^{tT} \sum\nolimits_{i\in I^t}{\left(
  \Sigma_i+{\bf q}_i{\bf q}_i^T\right)}{\bf w}^t$, where ${\bf
  q}_i={\bf m}_i-\hat{\bf m}_{I^t}$. The update rules for the
parameters are $\theta_{i1}=\tilde\theta_{i1}+\lambda\delta{\bf w}^t$,
$\theta_{i2}=\tilde\theta_{i2}+\lambda{\bf w}^t{\bf w}^{tT}$,
$\mu_{i2}=\tilde\Sigma_i-\lambda{\bf g}_i{\bf
  g}_i^T/\left(1+\lambda {\bf w}^{tT}{\bf g}_i\right)$, and
$\mu_{i1}=\Sigma_i\theta_{i1}$. 
The Woodbury Matrix Identity is used to avoid explicit matrix inversion in the computation
of $\mu_{i2}$. Again, solving for $v^t=\hat v^t$ gives an equation

\begin{small}
\begin{equation}
  \phi(\lambda)=\sum\nolimits_{i\in I^t}{\left(
    \Lambda_ic_i^2-f_i^2c_i^2+2f_ic_i(\delta-e_i)
    \right)}+\hat v^t-\tilde v^t=0,
\end{equation}
\end{small}

\noindent where $\phi(\lambda)$ is a monotone function, whose root can be
determined efficiently with a one-dimensional root finding algorithm. Notice
that $\Lambda_i$ and $f_i$ are functions of $\lambda$.

\begin{figure*}[tp]
\begin{tabular}{@{}ccc@{}}
\includegraphics[width=0.32\textwidth]{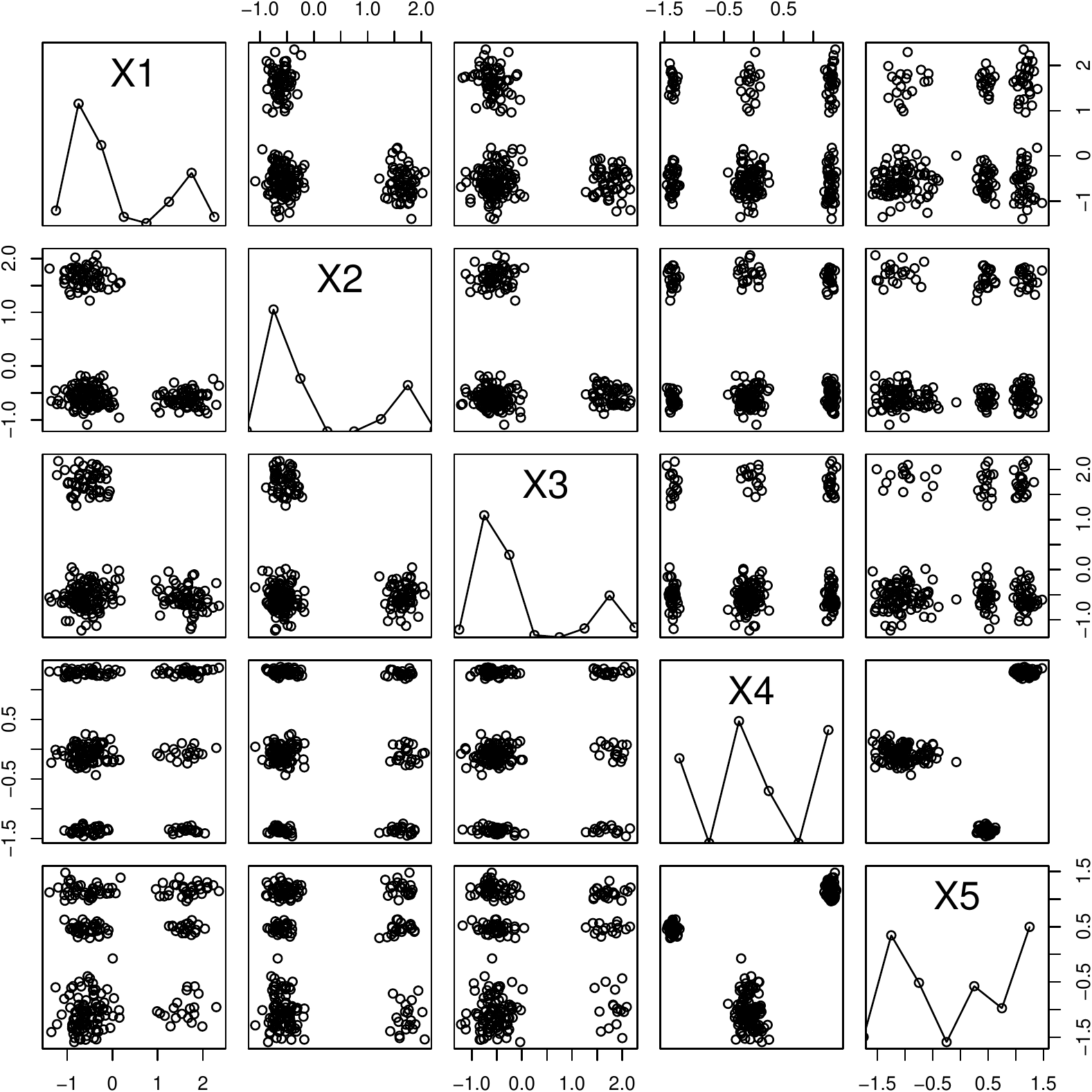} &
\includegraphics[width=0.32\textwidth]{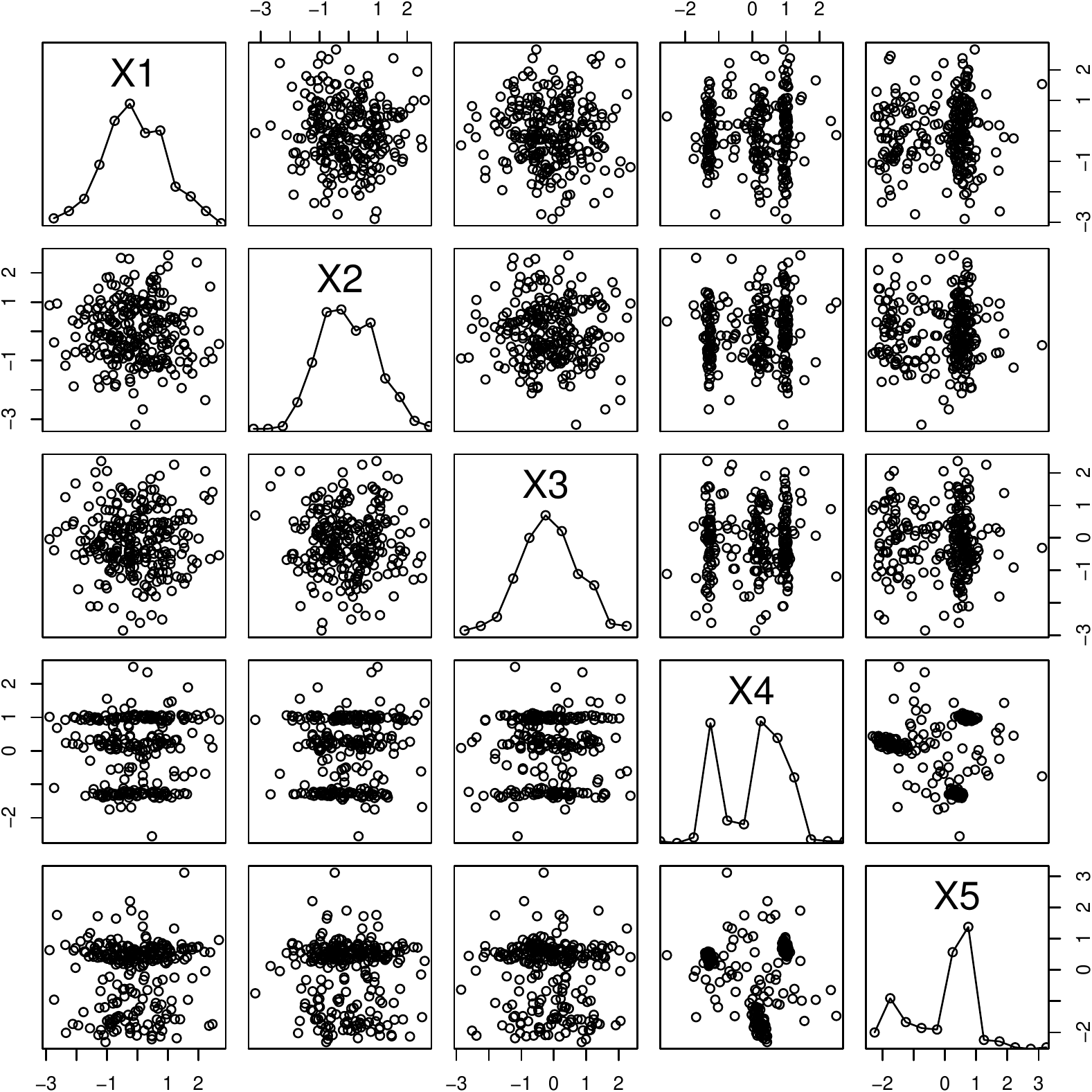} &
\includegraphics[width=0.32\textwidth]{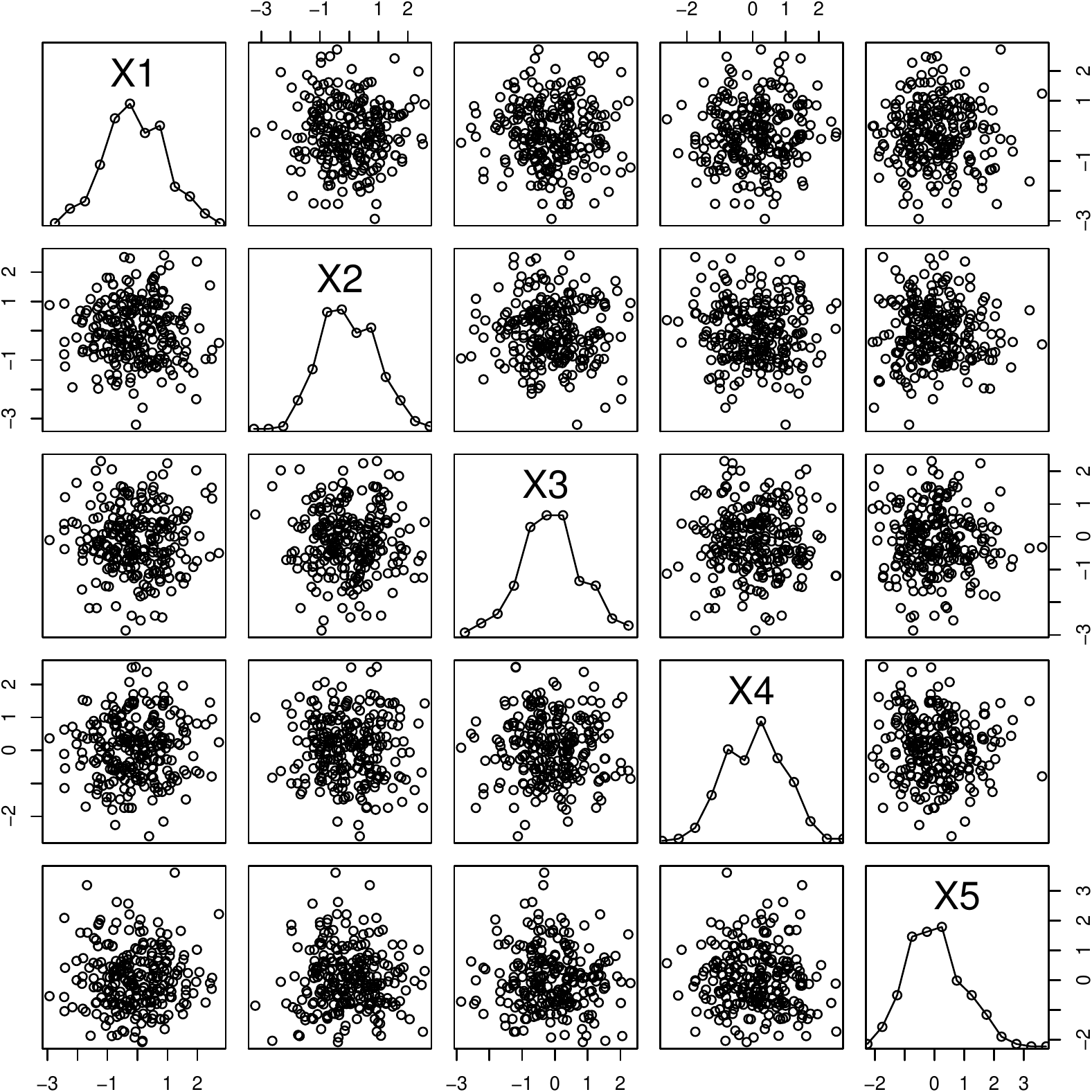}
\\
\small{(a)} & \small{(b)} & \small{(c)}
\end{tabular}
\caption{A pairplot of the whitened data  $\hat{\bf Y}_{5}$.
(a) Initially, i.e.,  without constraints,   $\hat{\bf Y}_{5} =\hat{\bf X}_{5}$.
(b) After the  cluster constraints are added for the four  clusters visible in Fig. \ref{fig:toy5_ica0}a.
(c) After further cluster constraints are added for the three  clusters visible in Fig. \ref{fig:toy5_ica0}c.
  \label{fig:toy5_white}}
\end{figure*}

\subsubsection{About convergence}

In the runtime experiment (Table \ref{tab:runtime}) we define the
optimization to be converged when the maximal absolute change in the
lambda parameters is $10^{-2}$ or when the maximal change in the means
or square roots of variance constraints is at most $10^{-2}$ times the
standard deviation of the full data. In the {\sc sideR} implementation
we stop the optimization after c. 10 seconds even if the above
conditions are not met. We describe in this section a situation where
the time cutoff is typically needed.
The iteration is guaranteed to converge eventually, but in certain
cases---especially if the dataset ($n$) is small or the size of some
clusters ($|I^t|$) is small compared to dimensionality ($d$)---the
convergence can be slow, as shown in the following adversarial
example.

Consider a dataset of three points ($n=3$) and a dimensionality of two
($d=2$), shown in Fig. \ref{fig:conv}a and given by
\begin{equation}
  \hat{\bf X}=\left(\begin{array}{cc}1&0\\0&1\\0&0\end{array}\right),
\end{equation}
and two sets of constraints. ({\sc A}) The first set of constraints
are the cluster constraints related to the first and the third row and
given by ${\mathcal C}_A=\{C^1,\ldots,C^4\}$, where $c^1=c^3=lin$,
$c^2=c^4=quad$, $I^1=\ldots=I^4=\{1,3\}$, $w^1=w^2=(1,0)^T$, and
$w^3=w^4=(0,1)^T$§. ({\sc B}) The second set of constraints have an
additional cluster constraint related to the second and the third row and
given by ${\mathcal C}_B=\{C^1,\ldots,C^8\}$, where $C^1,\ldots,C^4$ are as above
and $c^5=c^7=lin$, $c^6=c^8=quad$, $I^5=\ldots=I^8=\{2,3\}$,
$w^5=w^6=(1,0)^T$, and $w^7=w^8=(0,1)^T$.

(Case {\sc A}) The solution to Prob. \ref{pro:maxent} with
constraints in ${\mathcal C}_A$ is given by ${\bf m}_1={\bf m}_3=(1/2,0)^T$,
${\bf m}_2=(0,0)^T$,
\begin{equation}
  \Sigma_1=\Sigma_3=\left(\begin{array}{cc}1/4&0\\0&0\end{array}\right),~~~
    \mbox{ and } \Sigma_2=\left(\begin{array}{cc}1&0\\0&1\end{array}\right).
\end{equation}

Note that if the number of data points in a cluster
constraint is at most the number of dimensions in the data there are
necessarily directions in which the variance of the background
distribution is zero, see Fig. \ref{fig:conv}a. However, since we
have here a single cluster constraint with no overlapping
constraints, the convergence is very fast and in fact occurs after one
pass over the lambda variables as shown in Fig. \ref{fig:conv}b (black
line).

(Case {\sc B}) The solution to Prob. \ref{pro:maxent} with constraints in
${\mathcal C}_B$ are given by ${\bf m}_1=(1,0)^T$, ${\bf m}_2=(0,1)^T$, ${\bf
  m}_3=(0,0)^T$, and
\begin{equation}
  \Sigma_1=\Sigma_2=\Sigma_3=\left(\begin{array}{cc}0&0\\0&0\end{array}\right).\end{equation}
Here we observe that adding a second overlapping cluster
constraint, combined with the small variance directions in both the
constraints restricts the variance of the third data point to
zero. Because both of the clusters have only one additional data point
it follows that the variance of all data points is zero. The small variance and overlapping data points
cause the convergence here to be substantially slower, as shown in
Fig. \ref{fig:conv}b (red line). The variance scales roughly as
$(\Sigma_1)_{11}\propto\tau^{-1}$, where $\tau$ is the number of
optimization steps, the global optimum being in singular point at
$(\Sigma_1)_{11}=0$.

The slow convergence is due to overlapping and quadratic constraints
with small variance (caused here by small number of points per
cluster). A way to speed up the convergence would be---perhaps
unintuitively---to add more data points: e.g., to replicate
each data point 10 times with random noise added to each
replicate. When a data point would be selected to a constraint then
all of its replicates would be included as well. This would set a
lower limit on the variance of the background model and hence,
be expected to speed up the convergence. Another way to solve
the issue is just to cut off the iterations after some time
point---which is what we do here---leading up to larger
variance than in the optimal solution. This appears to be
typically acceptable in practice.

\subsection{Whitening out the background distribution}

Once we have found the distribution that solves
Prob.~\ref{pro:maxent}, the next task is to find and visualize the
maximal differences between the data and the background distribution
defined by Eq. \eqref{eq:maxent}. To this end we sample a dataset from
the background distribution, and produce a {\em whitened} version
of the data.

The random dataset can be obtained by sampling a data point $i\in[n]$
from the multivariate Gaussian distribution parametrized by
$\theta_i$.
The whitening operation is more involved. 
The direction-preserving whitening transformation of the data
results in a unit Gaussian spherical distribution, if the data follows
the current background distribution. Thus, any deviation from the unit
sphere distribution is a signal of difference between the data and the
current background distribution.

More specifically, we define new data vectors $\mathbf{ y}_i$ as
follows,
\begin{equation}\label{eq:y}
  \mathbf{ y}_i=U_iD_i^{1/2}U_i^T\left(\mathbf{ x}_i-\mathbf{
    m}_i\right),
\end{equation}
where the SVD decomposition of $\Sigma_i^{-1}$ is given by
$\Sigma_i^{-1}=U_iD_iU^T_i$, where $U_i$ is an orthogonal matrix and
$D_i$ is a diagonal matrix. If we used one transformation matrix for
the whole data, this would correspond to the normal whitening
transformation. However, here we may have a different transformation
for each of the rows. Furthermore, normally the transformation matrix
would be computed from the data, but here we compute it from the
constrained model.

\begin{figure*}[tp]
\centering
\includegraphics[width=0.95\textwidth]{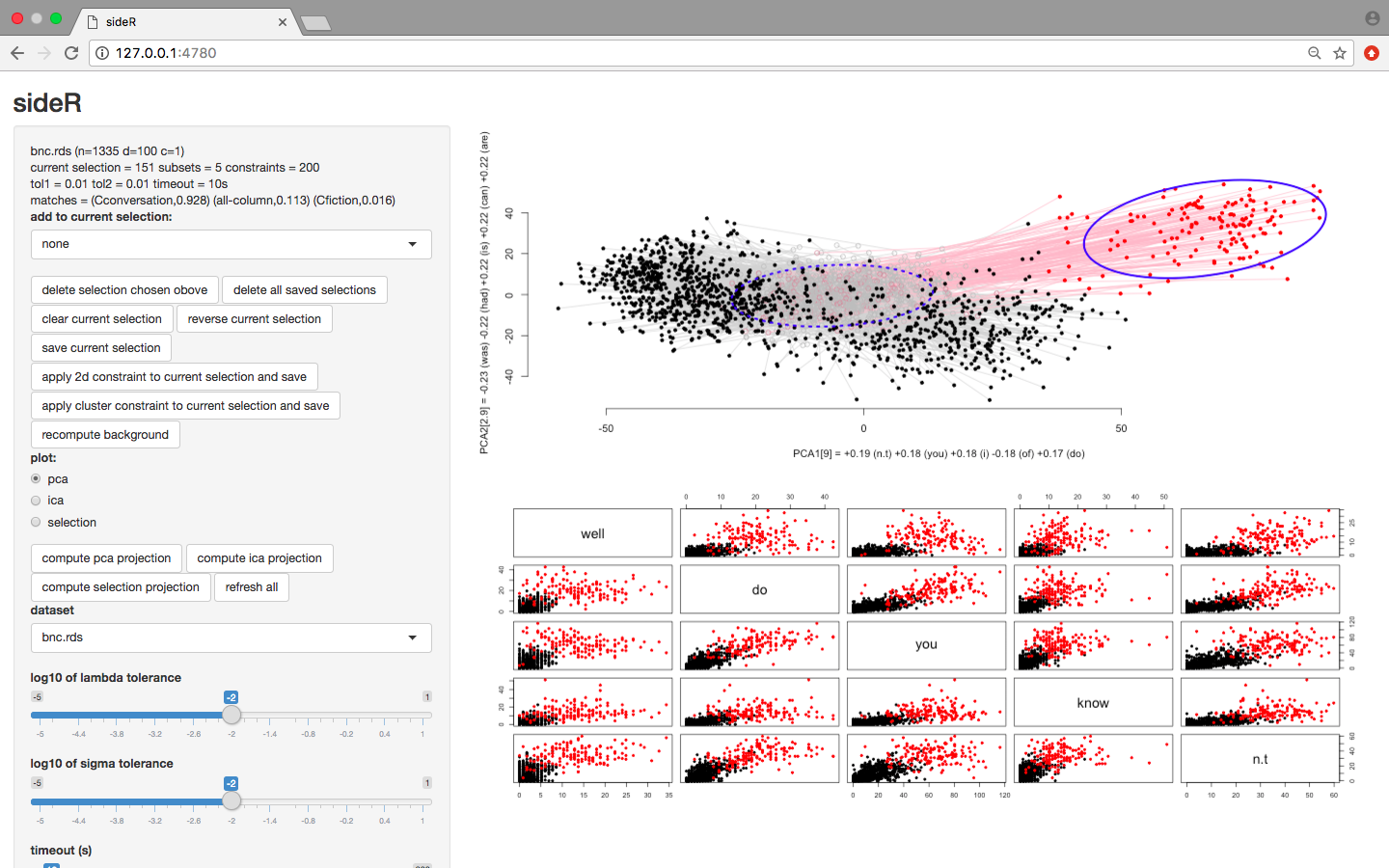}
\caption{\label{fig:sideRoverview}The full user interface of {\sc
    sideR}. The data shown here is the British National Corpus data, see Sec. \ref{sec:bnc} for details.}
\end{figure*}

It is easy to see that if $\mathbf{ x}_i$ obeys the distribution of
Eq. \eqref{eq:maxent}, then $D_i^{1/2}U_i^T\left(\mathbf{
  x}_i-\mathbf{m}_i\right)$ obeys unit spherical distribution. Hence,
any rotation of this vector obeys a unit sphere distribution as well. We
rotate this vector back to the direction of $x_i$ so that after the
final rotation, $\mathbf{ y}_i$ for different rows $i$ have a
comparable direction.  We use $\hat{\bf Y}= \left(\hat{\bf
  y}_1\hat{\bf y}_2\ldots\hat{\bf y}_n\right)^T$ to denote the whitened
data matrix. When there are no constraints ${\bf m}_i={\bf 0}$ and
$\Sigma_i^{-1}={\bf 1}$ and the whitening operation therefore reduces
to identity operation, i.e., $\hat{\bf Y}=\hat{\bf X}$.

\begin{example}
To illustrate the whitening operation, we show in Fig. \ref{fig:toy5_white}
pairplots of the whitened data matrix $\hat{\bf Y}_5$, for
the synthetic data $\hat{\bf X}_5$. Initially (Fig. \ref{fig:toy5_white}a)  the whitened data matches $\hat{\bf X}_5$.
Fig. \ref{fig:toy5_white}b shows the whitened data
after adding a cluster constraint
for each of the four clusters 
in Fig.~\ref{fig:toy5_ica0}a and
updating the background distribution accordingly. 
Now, in the first three dimensions the whitened data does not anymore significantly
differ from Gaussian distribution, while in dimensions 4 and 5 it still
does. 
\end{example}

\subsection{PCA and ICA}

To find directions where the data and the background distribution differ,
i.e., the whitened data $\hat{\bf Y}$ looks different from the unit
Gaussian distribution with zero mean, an obvious choice is to use
Principal Component Analysis (PCA) and look for directions in which
the variance differs most from unity.\footnote{Here we measure the
  difference of variance from unity to the direction of each principal
  component by $(\sigma^2-\log{\sigma^2}-1)/2$, where
  $\sigma^2$ is the variance to the direction of the principal
  component, and show in the scatterplot the two principal components
  with the largest differences from unity.} However, it may happen
that the variance is already taken into account in the variance constraints,
in which case PCA
is not informative because all directions in $\hat{\bf Y}$ have equal mean
and variance. Instead, we can for example use Independent Component
Analysis (ICA) and the FastICA algorithm \cite{hyvarinen1999fast} with
log-cosh $G$ function as a default method to
find non-Gaussian directions. Clearly, when there are no constraints, the
approach equals standard PCA and ICA on the original data but when
there are constraints the output will be different.

\begin{example}
Fig. \ref{fig:toy5_ica0}c shows the directions in which the whitened data $\hat{\bf Y}_5$ in Fig. \ref{fig:toy5_white}b differs the most from Gaussian. The user can  observe the cluster structure in dimensions 4 and 5, which would not be possible with non-iterative methods. After adding a cluster constraint for each of the three visible clusters, the updated
background distribution becomes a faithful representation of the data, and thus the whitened
data shown in Fig. \ref{fig:toy5_white}c  resembles a unit Gaussian spherical
distribution.
This is also reflected in a visible drop in ICA scores in Table \ref{tab:icascores}.
\end{example}

\section{Proof-of-concept system {\sc sideR}}
\label{sec:sider}

We have implemented the concepts described above in an interactive
demonstrator system {\sc sideR}
using R 3.4.0 \cite{R} with {\sc shiny} \cite{Rshiny} and {\sc fastICA}
\cite{RfastICA}.
{\sc sideR} runs in the web browser using R as a back-end.%
\footnote{  {\sc sideR}  is 
    released as a free open source system under the MIT license.
{\sc
    sideR} and the code used to run the convergence and runtime
  experiments (Fig. \ref{fig:conv} and Tab. \ref{tab:runtime}) is
  available for download at 
  \url{http://www.iki.fi/kaip/sider.html}.
     }
The user interface of {\sc sideR} is shown in
Fig. \ref{fig:sideRoverview}. The main scatterplot (upper right
corner) shows the PCA (here) or ICA projection of the data to
directions in which the data and the background distribution differ
most. The data is shown by solid black spheres and the currently
selected data by solid red spheres. The sample from the background
distribution is shown by gray circles, with a gray line connecting the
data points with the corresponding sampled background points. Notice
that the gray points and lines only provide a proxy for the difference
between the data and the background distribution, since in reality the
background distribution has a specified mean and covariance structure
for every point. Nonetheless this should illustrate broadly the density
structure of the background distribution for the current projection, which
we think is helpful to understand why the current visualization may
provide new insights.

We also show a pairplot (lower right corner) directly dislaying the attributes
maximally different with respect to the current
selection  (red points) as compared to the full dataset. In the
left-hand panel we show some statistics of the full data and of the
data points that have been selected. We also show in the main
scatterplot (upper right corner) in blue the 95\% confidence
ellipsoids for the distribution of the selection (solid blue ellipsoid)
and the respective background samples (dotted blue ellipsoid) to aid
in figuring out if the location of the selected points in the current
projection differs substantially from the expected location under the
background distribution.\footnote{The 95\% confidence regions are here
  a visual aid computed from the points shown in the projection. The
  confidence ellipsoid could be computed also from the background
  distribution directly, but it is a simplification as well since every data point may have unique mean and co-variance parameters.}

The user can add data points to a selection by directly marking them,
by using pre-defined classes that exist in the dataset, or previously
saved groupings. The user can create a 2-D or cluster constraint of the
current selection by clicking the appropriate button on the left-hand
panel as well as recompute the background distribution to match the
current constraints and update the projections. The user can also
adjust convergence parameters which have by default been set so that
the maximal time taken to update the background distribution is $\sim$10
seconds which is in practice typically more than enough. The interface
has been designed so that time-consuming operations (taking more than
$\sim$2 seconds, i.e., the updating the background distribution
or computing the ICA projection) are executed only by a direct command
by the user, which makes the system responsive and predictable.

\section{Experiments}
\label{sec:experiments}

In this section we demonstrate the use of  our method and the tool {\sc sideR}. Our focus here is to show how the tool is able  to  provide the user with insightful projections of data and reveal the differences between the background distribution and the data.
Additionally, the user interface makes it easy to explore
 various statistics and properties of  selections of data points.

We start by a runtime experiment in which we test the system with data
set sizes   typical for interactive systems, i.e., there are
on the order of thousands of data points; if there are more data points it
often makes sense to downsample the data first.
Then we use 
real datasets to illustrate how the system is able to find relevant projections for the user and display differences between the background distribution and the data.

\subsection{Runtime experiment}

We generated synthetic datasets parametrized by the
number of data points ($n$), dimensionality of the data ($d$), and the
number of clusters ($k$).  Each dataset was created by first randomly
sampling $k$ cluster centroids and then allocating data points around
each of the centroids.  We added column constraints ($2d$ constraints)
for each dataset and for the datasets with $k>1$ we additionally used
cluster constraints for each of the $k$ clusters in the data ($2dk$
constraints).  In Table~\ref{tab:runtime} the median wall clock running
times are provided without any cut-off, based on 10 runs for each set of
parameters, ran on a Apple MacBook Air with 2.2 GHz Intel Core i7
processor and a single-threaded R 3.4.0 implementation of the algorithm.

The algorithm is first initialized ({\sc init}) which is typically
very fast, after which the correct parameters are found ({\sc
  optim}). Then preprocessing is done for sampling and whitening ({\sc
  preprocess}) after which we produce a whitened dataset ({\sc
  whitening}) and a random sample of the maxent distribution ({\sc
  sample}). These are then used to run the PCA ({\sc pca}) and ICA ({\sc
  ica}) algorithms. We found that {\sc init}, {\sc preprocess}, {\sc whitening},
{\sc sample}, and {\sc pca} always take less than 2 seconds each and
they are not reported in the table. Most of the time is consumed by
{\sc optim}. We observe in Table~\ref{tab:runtime} that, as expected,
the time consumed does not depend on the
number of rows $n$ in the dataset. Each of the optimization steps
takes $O(d^2)$ time per constraint and there are $O(kd)$ constraints,
hence the time consumed scales as expected roughly as $O(kd^3)$.
In {\sc sideR} the default setting is to stop the optimization after a time
cut-off of 10 seconds, even when convergence has not been achieved.
For larger matrices the time consumed by ICA becomes significant,
scaling roughly as $O(nd^2)$.

\begin{table}[t]
\centering
    \caption{\label{tab:runtime} Median wall clock running times, based
      on 10 runs for each set of parameters for finding the correct
      parameters ({\sc optim}) and running the ICA ({\sc ica})
      algorithm without time cutoff.}
    \begin{footnotesize}
\begin{tabular}{cc|cc}
\multicolumn{2}{c}{}&\multicolumn{2}{c}{seconds, $k\in\{1,2,4,8\}$}\\
$n$&$d$&{\sc optim}&{\sc ica}\\\hline
$2048$&$16$&$\{0.0,0.2,0.3,0.5\}$&$\{0.6,0.6,0.6,0.6\}$\\
$2048$&$32$&$\{0.0,0.6,1.0,2.1\}$&$\{1.5,1.5,1.6,1.6\}$\\
$2048$&$64$&$\{0.1,2.7,5.2,11.0\}$&$\{5.1,5.2,4.9,4.9\}$\\
$2048$&$128$&$\{1.2,21.4,48.1,124.6\}$&$\{17.8,17.6,17.4,17.0\}$\\
$4096$&$16$&$\{0.0,0.2,0.3,0.5\}$&$\{1.1,1.1,1.1,1.1\}$\\
$4096$&$32$&$\{0.0,0.6,1.0,2.0\}$&$\{3.1,3.4,3.0,3.1\}$\\
$4096$&$64$&$\{0.2,2.5,6.0,11.6\}$&$\{9.8,9.3,9.5,9.6\}$\\
$4096$&$128$&$\{1.2,23.4,56.4,121.3\}$&$\{34.2,34.7,34.4,34.4\}$\\
$8192$&$16$&$\{0.0,0.2,0.3,0.6\}$&$\{2.6,2.2,2.5,2.1\}$\\
$8192$&$32$&$\{0.0,0.6,1.0,2.0\}$&$\{6.5,6.0,5.9,5.9\}$\\
$8192$&$64$&$\{0.2,2.7,6.0,12.2\}$&$\{20.7,20.4,19.8,20.1\}$\\
$8192$&$128$&$\{1.2,21.9,44.1,110.3\}$&$\{67.9,67.5,67.1,67.6\}$\\
\end{tabular}
\end{footnotesize}
  \end{table}

  \begin{figure*}[tp]
  \begin{center}
\begin{tabular}{cc}
\includegraphics[width=0.85\columnwidth]{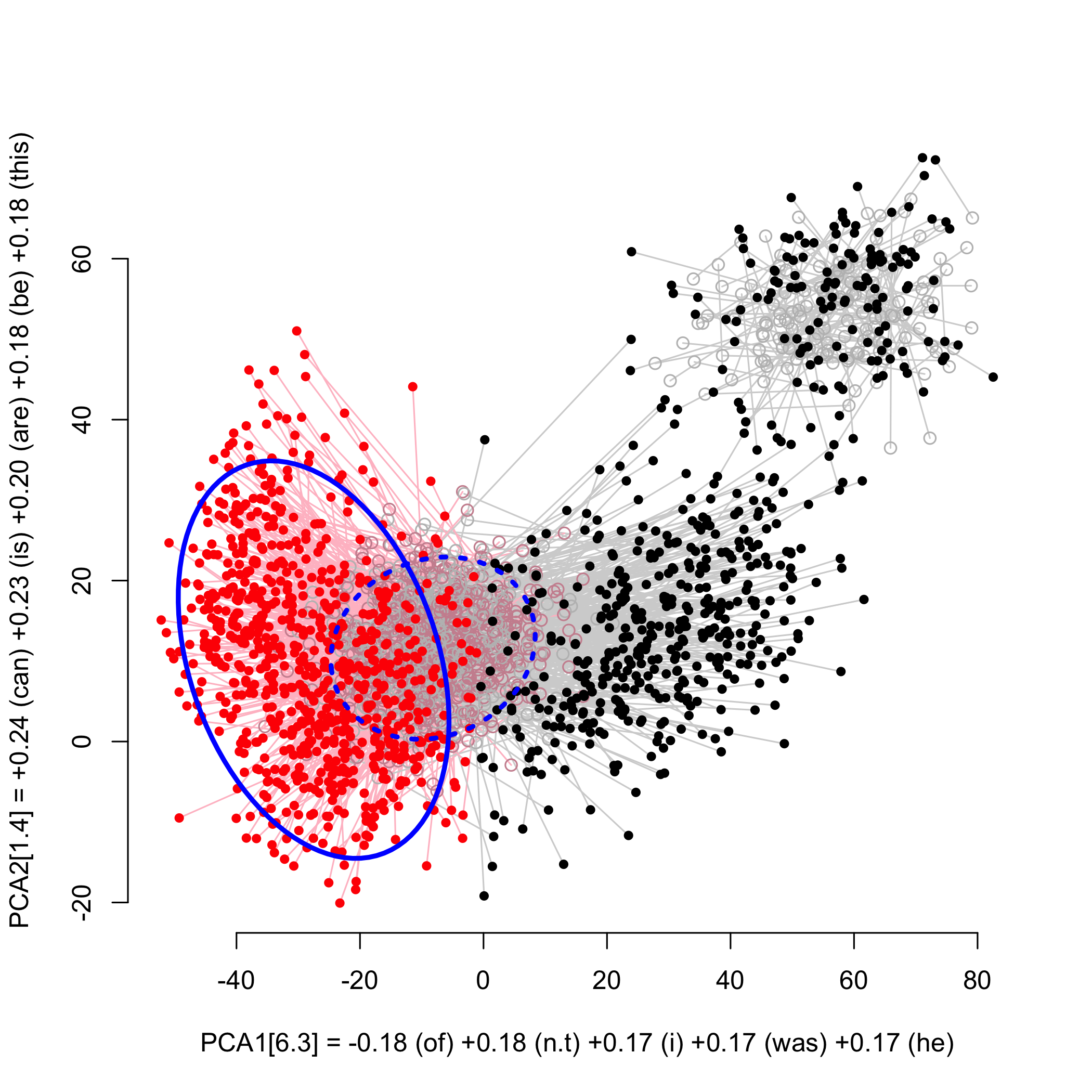} &
\includegraphics[width=0.85\columnwidth]{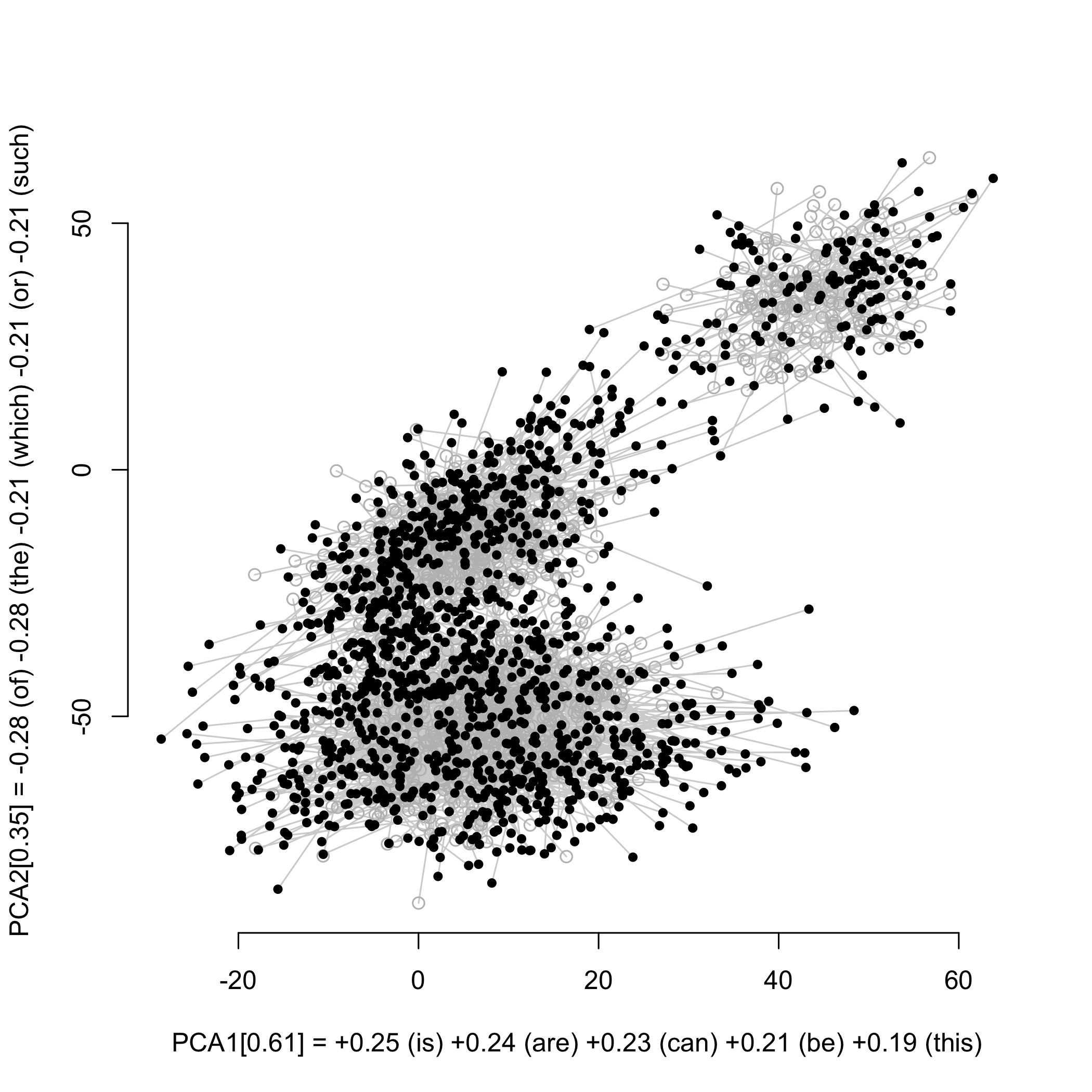} \\
\small{(a)}  & \small{(b)}
\end{tabular}
\end{center}
\caption{A use case with the BNC data.
(a) Selection of points for the second cluster constraint. The view is the next most informative PCA projection
obtained after adding a cluster constraint for the points selected in Fig. \ref{fig:sideRoverview} and updating of the background distribution.
(b)  After adding a cluster constraint again and updating of the background distribution, there is no longer a striking difference between the background distribution and the data in the PCA projection.}
\label{fig:bnc}
\end{figure*}

\subsection{British National Corpus data}
\label{sec:bnc}

The British National Corpus (BNC) \cite{bnc:2007} is one of the largest annotated text corpora freely available in full-text format. The texts are  annotated with information such as author gender, age, and target audience, and all texts have been classified into genres \cite{lee:2001}.
%
%
As a  high dimensional use case we explore the high-level structure of the corpus. For  a preprocessing step, we compute the vector-space model (word counts) using the first 2000 words from each text belonging  to one the four main genres  in the corpus (`prose fiction', `transcribed conversations', `broadsheet newspaper', `academic prose') as done in \cite{lijffijtForthcoming}.
After preprocessing we have word counts for 1335 texts and we use the 100 words with highest counts as the dimensions and the main genres as the class information.

The most informative PCA projection to the BNC data is shown in Fig. \ref{fig:sideRoverview}.
In the upper right corner there is a group of points (red selection) that appear to form a group.  The selected points are mainly texts from `transcribed conversations' (Jaccard-index to class 0.928), and the pairplot in  Fig. \ref{fig:sideRoverview} (lower right) shows  the selection of points differing from the rest of the data. After we added a cluster constraint for this selection, updated the background distribution and computed a new PCA projection, we obtained the projection in Fig. \ref{fig:bnc}a.

The next selection shows another set of points differing from the background distribution. This set of points mainly contains `academic prose' and  `broadsheet newspaper' (Jaccard-indices 0.63 and 0.35).
After adding a cluster constraint for this selection, we updated the background distribution and computed another PCA projection, resulting in the projection in Fig. \ref{fig:bnc}b. Now, there is no apparent difference to the background distribution (reflected indeed in low PCA scores), and we conclude that
the identified `prose fiction' class, together with the combined 
cluster of `academic prose' and  `broadsheet newspaper'
 explain the data well wrt. variation in counts of the most frequent words. Notice that we did not provide the class labels in advance, they were only used retrospectively.

 \begin{figure*}[tp]
\begin{tabular}{@{}c@{}c@{}c@{}}
\includegraphics[width=0.33\textwidth]{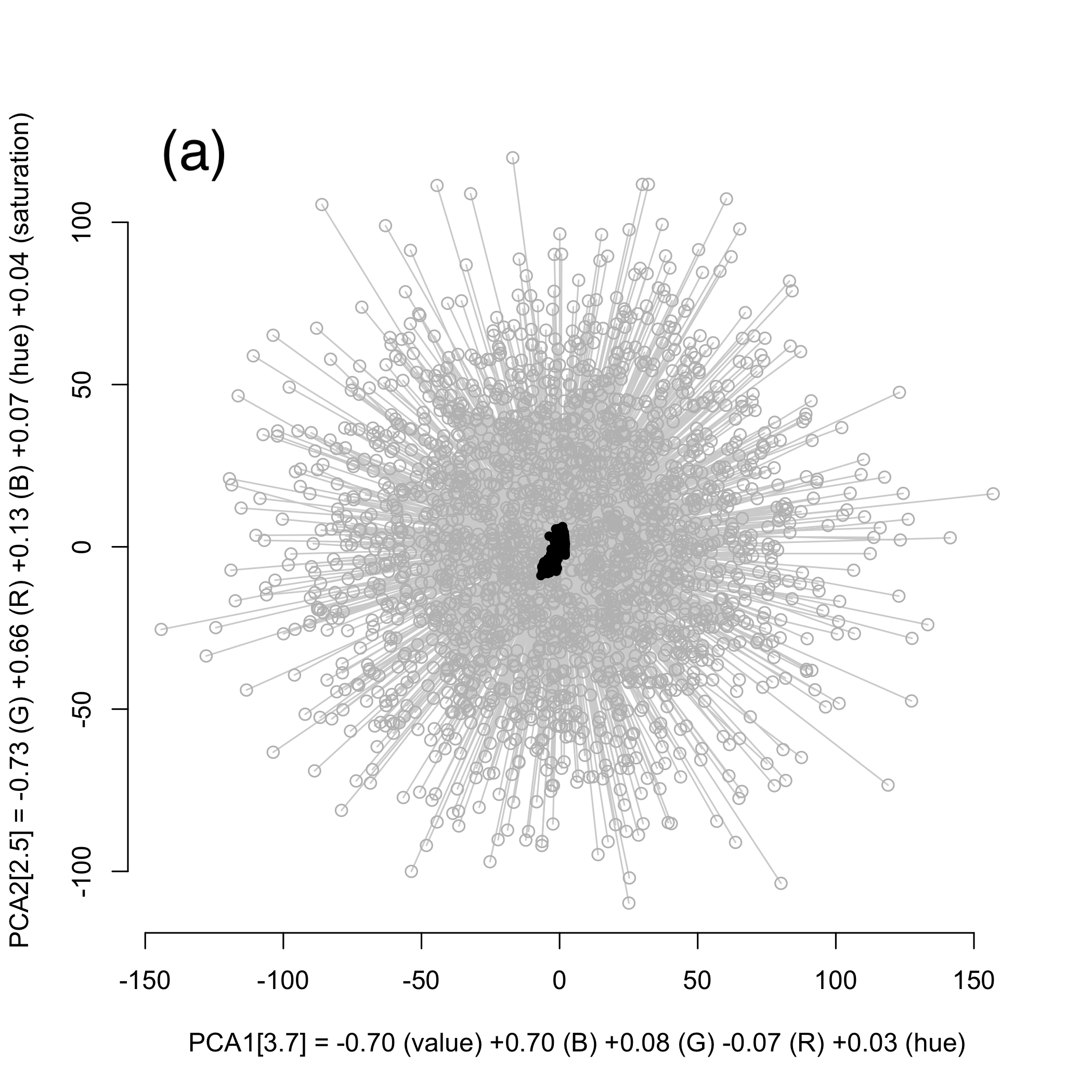} &
\includegraphics[width=0.33\textwidth]{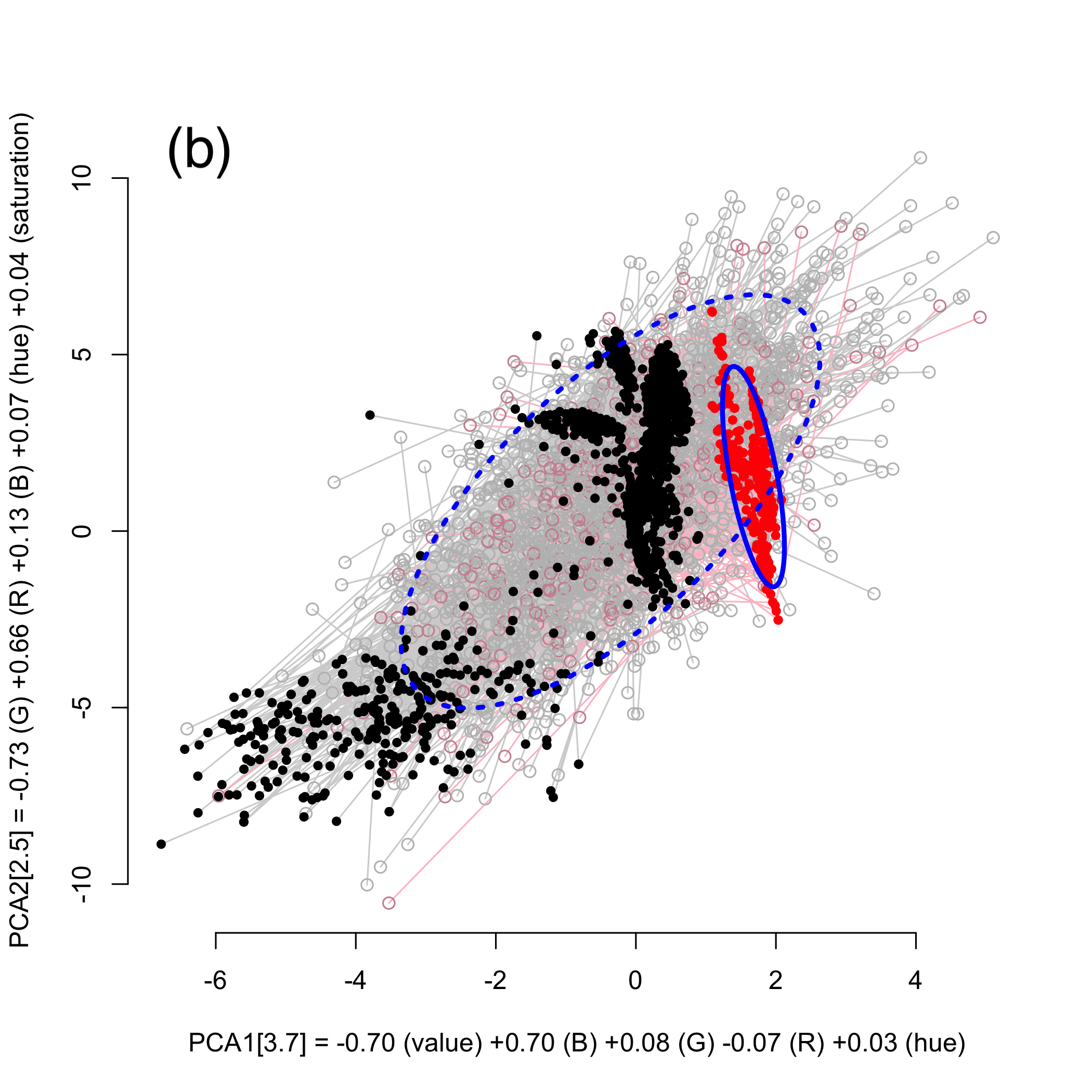} &
\includegraphics[width=0.33\textwidth]{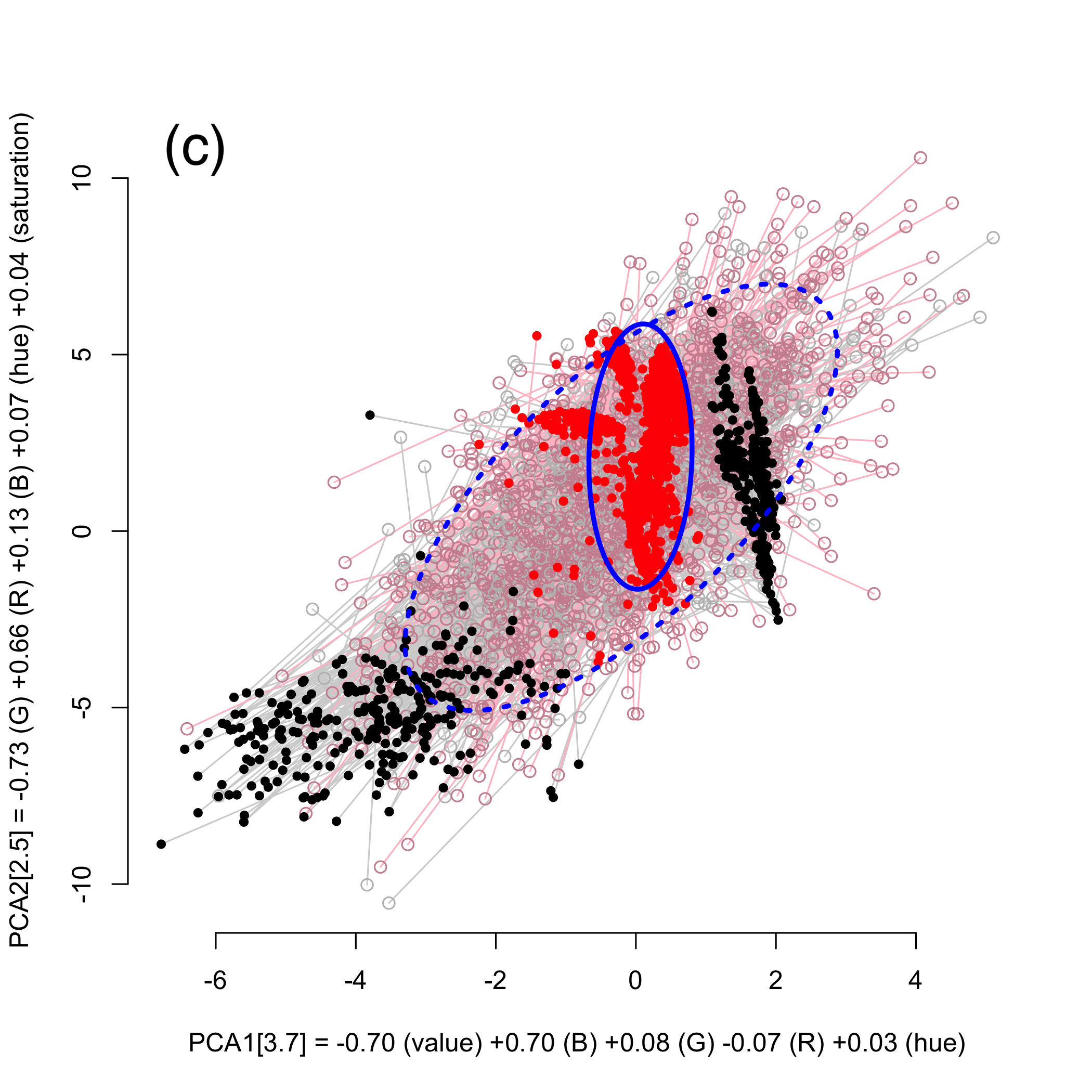} \\
\includegraphics[width=0.33\textwidth]{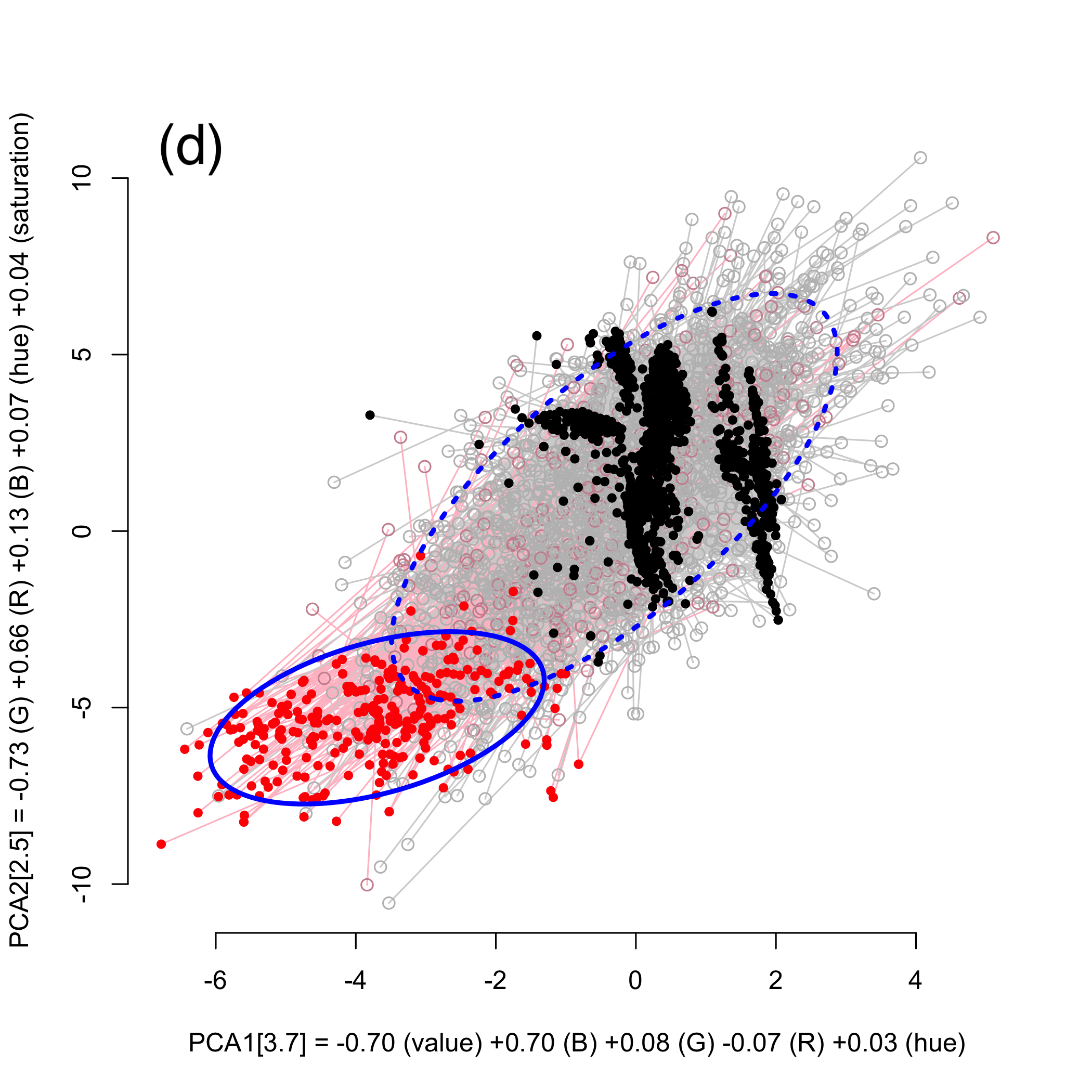} &
\includegraphics[width=0.33\textwidth]{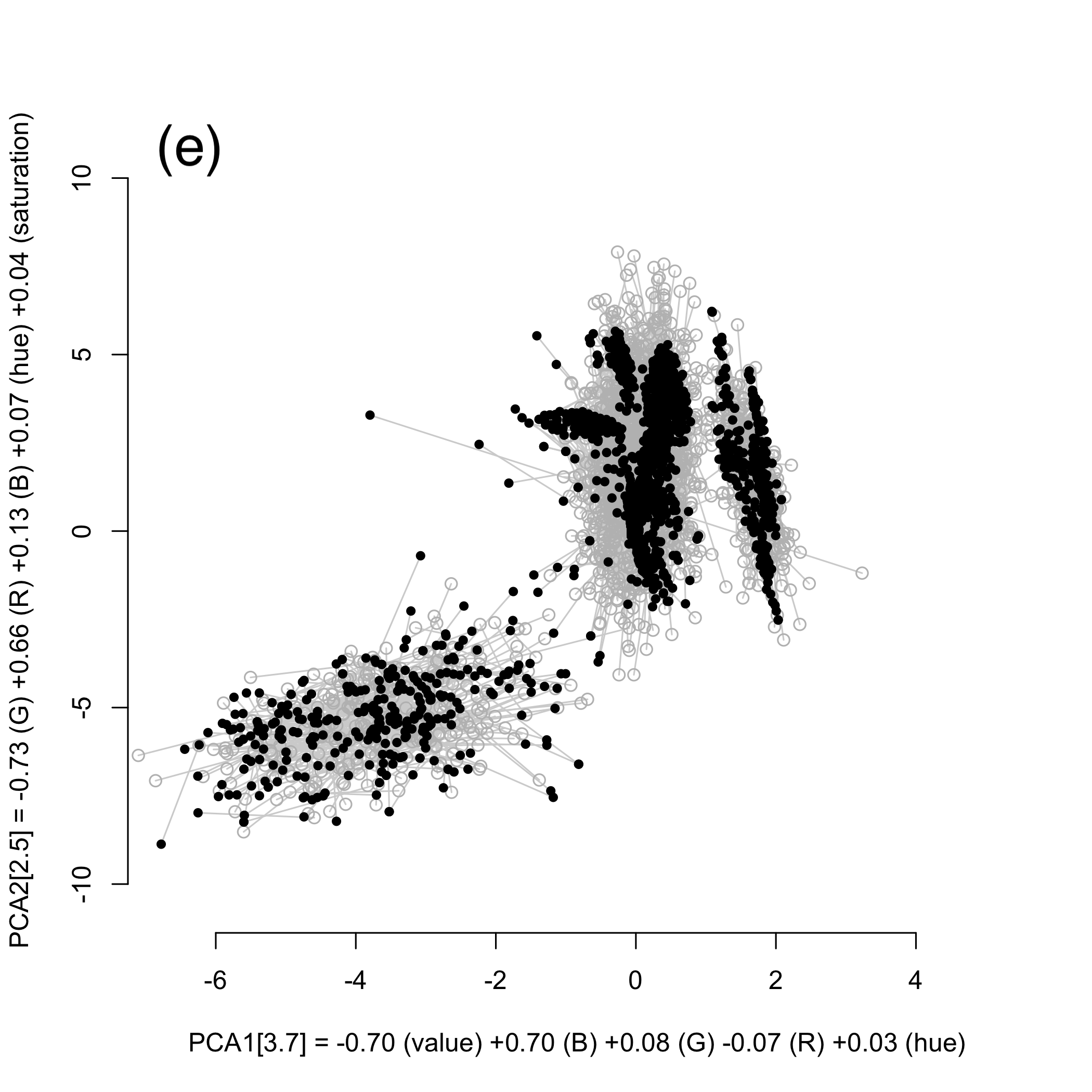} &
\includegraphics[width=0.33\textwidth]{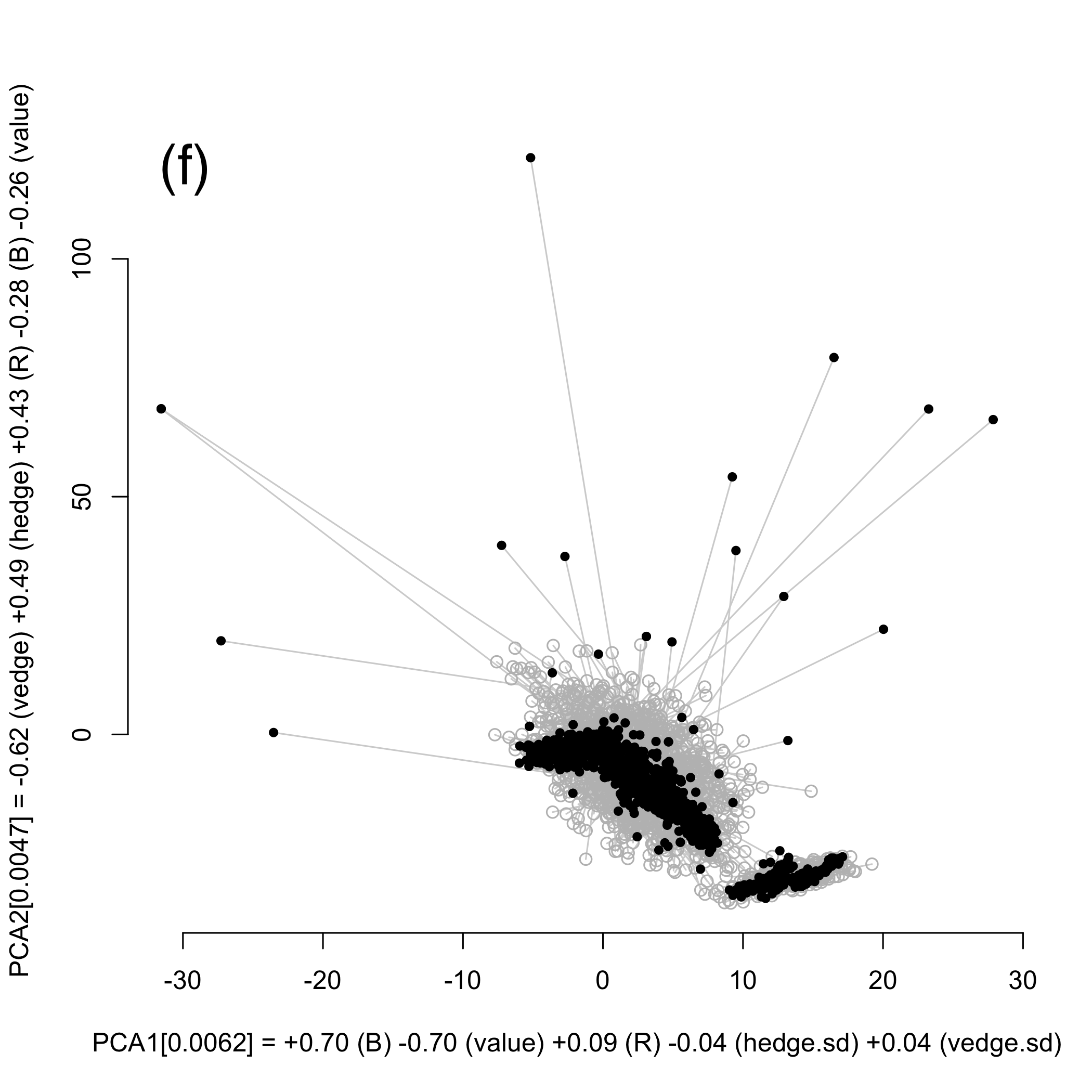}
\end{tabular}
\caption{A use case with the UCI Image Segmentation data.
(a) The first view  shows that initially the scale of  background distribution significantly differs from that of the data.
(b) After adding a 1-cluster constraint and performing an update of the background distribution there is visible structure present. The points selected for the first cluster constraint are shown in red. (c) and (d) The selections of points for the second and third cluster constraint, respectively.
(e) After three cluster constraints are added and  the background distribution is updated accordingly, the data and the background distribution are similar in this projection.
(f) The next PCA projection shows mainly outliers.}
\label{fig:segment}
\end{figure*}

\subsection{UCI Image Segmentation data}

As a second use case, we have the Image Segmentation dataset
from the UCI machine learning repository \cite{Lichman:2013} with
2310 samples. The PCA projection (Fig. \ref{fig:segment}a) shows that the background distribution has a much larger variance than the data. Thus, we first added a 1-cluster constraint for the data (overall covariance) and updated the background distribution. After this (Fig. \ref{fig:segment}b) $\geq 3$ sets of points quite clearly separated in the projection can be observed. The set of 330 points selected in Fig. \ref{fig:segment}b contains solely points from the class `sky', while the 316 points in the lower left corner (selected in Fig. \ref{fig:segment}d) are mainly from the class `grass' (with Jaccard-index 0.964). The set of points clustered in the middle (selected in Fig. \ref{fig:segment}c) are mainly   from classes `brickface', `cement', `foliage', `path', and `window' (with Jaccard-index approx. 0.2 each).
We add a cluster constraint for each of selection, and show the
updated background distribution in
Fig. \ref{fig:segment}e. We can observe that the background
distribution now matches the data rather well with the exception of
some outliers. The next PCA projection 
(Fig. \ref{fig:segment}f) reveals that indeed there are outliers. For
brevity, we did not continue the analysis, but the data obviously
contains a lot more structure that we could explore in subsequent
iterations. Furthermore, in many applications identifying and studying
the outlier points deviating from the three-cluster main structure
of the data could be interesting and useful.

\section{Related Work}\label{sec:related}

This work is motivated by the ideas in \cite{puolamaki2016ecmlpkdd} in
which a similar system was constructed using constrained
randomization. The constrained randomization approach
\cite{hanhijarvi2009,lijffijt2014} is similar to the Maximum Entropy
distribution used here, but it relies on sampling of the data and no
direct distribution assumptions are made. An advantage of the approach
taken here is that it is faster---which is essential in interactive
applications---and scales more easily to large data. Furthermore, here
we have an explicit analytic form for the background distribution
unlike in \cite{puolamaki2016ecmlpkdd}, where the background
distribution was defined by a permutation operation. 
The mathematical form of linear and quadratic constraints and efficient inference of 
the background distribution has been developed by us in \cite{lijffijt_forthcoming}.
The presentation here is new and non-overlapping.
The 
analytic form of the background distribution allows us, in
addition to speeding up the computations, to define
interestingness functions and the cluster constraint in a more natural
manner. Here we also introduce the whitening method that allows our
approach to be used with standard and robust projection pursuit
methods such as PCA or ICA instead of the tailor-made line search
algorithm of \cite{puolamaki2016ecmlpkdd}. Furthermore, we provide a
fluent open source implementation written in R.

The Maximum Entropy method has been proposed as a part of Formalizing
Subjective Interestingess ({\sc forsied}) framework of data mining
\cite{debie2011,debie2013} modelling the user's knowledge  by a
background distribution. {\sc forsied} has been studied in the context of
dimensionality reduction and EDA \cite{DLS:16,kang2016}. 
To the best of our knowledge, ours is the first instance in which
 this background distribution can be updated by a direct interaction of the user, 
 thus providing a principled method of EDA.

Many other special-purpose methods have been developed for
active learning 
in diverse settings, e.g., in 
classification and ranking, as well as explicit models for user
preferences. However, as these approaches are not targeted at data
exploration, we do not review them here. Finally, several special-purpose
methods have been developed for visual iterative data exploration in specific
contexts, e.g., for itemset mining and subgroup discovery
\cite{boley2013,dzyuba2013,vanleeuwen2015,paurat2014}, information retrieval
\cite{ruotsalo2015}, and network analysis \cite{chau2011}.

The system
presented here can be also considered to be an instance of {\em
  visually controllable data mining} \cite{puolamaki2010}, where the
objective is to implement advanced data analysis methods 
understandable and efficiently controllable by the user. Our
approach satisfies the properties of a visually controllable
data mining method (see \cite{puolamaki2010}, Sec. II B): (VC1) the
data and model space are presented visually, (VC2) there are intuitive
visual interactions allowing the user to modify the model space, and
(VC3) the method is fast enough 
for visual interaction.

Dimensionality reduction for EDA has been studied for decades
starting with
multidimensional scaling (MDS) \cite{kruskal1964,torgerson1952} and
projection pursuit \cite{friedman1974,huber1985}. Recent research
on this topic (referred to as manifold learning) is still
inspired by 
MDS: find a
low-dimensional embedding of points representing well the 
 the distances in the
high-dimensional space. In contrast to PCA
\cite{pearson1901}, 
the idea is to preserve small distances,
and large distances are irrelevant, as long as they remain large, e.g.,
Local Linear and (t-)Stochastic Neighbor Embedding
\cite{hinton2002,roweis2000,vandermaaten2008}. This is typically
not possible to achieve perfectly, and a trade-off between precision
and recall arises~\cite{venna2010}. 

Many new interactive visualization methods are presented yearly at the IEEE VIS Conference. The focus there is not on the use or development of new data mining or machine learning techniques, but rather on human cognition, efficient use of displays, and exploration via data object and feature selection. Yet, there is a growing interest and potential synergies with data mining were already recognized long ago \cite{vismasterbook}.

\section{Conclusions}\label{sec:conclusion}

There have been many efforts in analysis of multivariate data in
different contexts. 
For example, there are several
Projection Pursuit and manifold learning methods for 
using some criteria to compress the data into a
lower dimensional---typically 2-D---presentation while preserving features of interest.
The inherent drawback of this approach is that the criteria
for dimensionality reduction are
defined typically in
advance and it may or may not fit the user's need. It may be that a
visualization shows only the most prominent features of the data 
already known for the user, or features that are irrelevant for
the task at hand.

The advantage of the dimensionality reduction
methods is that the computer, unlike the human user, has a ``view''
of all the data and it can select a view in a more fine-tuned way and
by using a more complex criteria than a human could.
A natural alternative to static visualizations using  pre-defined
criteria is the addition of interaction. The drawback of such
interactions is, however, that they lack the sheer computational
power utilized by the dimensionality reduction methods.

Our method fills the gap between automated dimensionality
reduction methods and interactive systems. We propose to model the
knowledge of a domain expert by a probability distribution 
computed by using the Maximum Entropy criteria. Furthermore, we
propose powerful and yet intuitive interactions  for the user to
update the background distribution. Our approach
uses Projection Pursuit methods and  shows the directions in which
the data and the background distribution differ the most. In this
way, we utilize the power of Projection Pursuit at the same the allowing the user to adjust the criteria by which the
computers chooses the directions to show her.


The current work presents a framework and a system for real-valued
data and the background distribution which is modeled by multivariate
Gaussian distributions. The same ideas could be generalized to other
data types, such as categorical or ordinal data values, or to higher-order statistics,
likely in a straightforward manner,
as the mathematics of exponential family distribution would lead to 
similar derivations.

Our approach could also be extended to 
other interactions especially in knowledge intensive tasks. Instead of
designing interactions directly and explicitly we can think  that
the ``views of the data'' (here mainly 2-D projections) and the
interactions (here, e.g.,  marking the constraints)  could
also in other contexts be modeled as  operations modifying the user's
``background model''.
One of the main benefits of our approach is that the user marks the
patterns she observes and thus the background distribution is
always customized  to the user's understanding of the data,
without a need for assumptions such as that high variance directions
are interesting to everyone, as implicitly assumed when
applying PCA for visualization.

For concrete applications for our approach and the {\sc sideR} tool there
is potential in, e.g., computational flow cytometry. Initial experiments with
samples up to tens of thousands rows from flow-cytometry data
\cite{saeys2016computational} has shown  the computations in
{\sc sideR} to scale up well and the projections to reveal structure in the
data potentially interesting to the application specialist.

\vspace{2mm}\noindent{\em Acknowledgements.} This work has been supported by the ERC under the EU's Seventh Framework Programme (FP/2007- 2013) / ERC Grant Agreement no. 615517, the FWO (project no. G091017N, G0F9816N), the EU's Horizon 2020 research and innovation programme and the FWO under the MSC Grant Agreement no. 665501, the Academy of Finland (decision 288814), and Tekes (Revolution of Knowledge Work project).

\enlargethispage{-13pt}

\bibliographystyle{IEEEtran}
\bibliography{paper}

\end{document}